\documentclass[12pt]{article}

\usepackage{newtxtext,newtxmath}

\usepackage{graphicx}
\usepackage{mathtools}
\usepackage{bm}
\usepackage{float}
\usepackage[letterpaper,margin=1in]{geometry}

\renewenvironment{abstract}
	{\quotation}
	{\endquotation}

\date{}

\makeatletter
\renewcommand{\fnum@figure}{\textbf{Figure \thefigure}}
\renewcommand{\fnum@table}{\textbf{Table \thetable}}
\makeatother

\usepackage{scicite}

\usepackage{url}

\def\scititle{
	Why We Look Where We Look: Emergent Human-like Fixations of a Foveated Visual Language Model Maximizing Scene Understanding
}
\title{\bfseries \boldmath \scititle}

\author{
	Shravan~Murlidaran$^{1\ast\dagger}$,
    Ziqi Wen$^{3}$,
    Sana Shehabi$^{1}$,
    Miguel P.~Eckstein$^{1,2,3\dagger}$\and
	\small$^{1}$Psychological \& Brain Sciences, University of California, Santa Barbara \& 93106, USA.\and
	\small$^{2}$Electrical and Computer Engineering, University of California, Santa Barbara \& 93106, USA.\and
    \small$^{3}$Computer Science, University of California, Santa Barbara \& 93106, USA.\and
	\small$^\ast$Corresponding author. Email: smurlidaran@ucsb.edu\and
	\small$^\dagger$These authors contributed equally to this work.
}

\begin{document} 

\maketitle

\begin{abstract} \bfseries \boldmath
When humans view scenes without a specific task (free-viewing), they initially direct their eye movements toward the scene center and then fixate on people, text, objects being gazed at or grasped, and semantically meaningful regions. What these signature fixation patterns reflect and whether they optimize an underlying perceptual task remain unknown. We show that a computational agent with simulated foveation, trained to optimize scene comprehension, exhibits emergent human fixation signature patterns. In contrast, versions of the agent trained to search or classify scenes, or equipped with peripheral vision that was better or worse than human vision, predicted human fixation patterns less accurately. Thus, human free-viewing fixation patterns may emerge as a functional byproduct of optimizing scene comprehension under the biological constraints of foveated vision.

\end{abstract}

\section*{Introduction}
Humans explore the visual world through eye movements, directing central vision (the fovea) to regions of interest to process them with greater spatial detail. Decades of research show that these rapid eye movements are guided to locations and visual attributes that help us attain maximum accuracy when searching for targets \cite{eckstein2001quantifying,  najemnik_optimal_2005, malcolm2010combining, hoppe2019multi}, identifying faces \cite{peterson2012looking}, or executing actions such as grasping and walking \cite{gegenfurtner2016interaction, de2021functional, hayhoe2022visual, rothkopf2025computational}. But where humans look when free-viewing a scene without a particular task, such as when leisurely sitting at an outdoor cafe, glancing from a bus window, or entering a room, is not well understood. The notion that during free-viewing humans look at low-level salient regions defined by contrast, color, and orientation has been repeatedly challenged over the last 20 years \cite{einhauser2008objects, koehler2014saliency, henderson2017meaning, peacock2019role, murlidaran2025eye}. There is converging evidence for a set of landmark signatures of human eye movement behavior during free-viewing. Humans look at other people \cite{cerf2009faces, birmingham2008gaze}, text\cite{wang2012attraction, de2019individual}, and objects\cite{einhauser2008objects, nuthmann2010object}. In particular, they look at objects judged meaningful, objects that are gazed at, and objects that are grasped \cite{land1999roles, castelhano2007see, peacock2019role}. And when they first look at an image, they tend to look at the center \cite{ruddock1996relationship, tatler2007central}, an effect known as the central bias. But how might the characteristic and pervasive human fixation behaviors arise?  Could these emerge as byproducts of optimizing the visual system for a single task or a small number of specific tasks? We will call such a theory the task-optimization of free-viewing eye movements. It asserts that where humans look emerges from learning to efficiently explore a scene to attain the highest possible accuracy in performing some type of visuo-cognitive task. The frequent free-viewing fixation patterns to people, text, and objects are what we might expect from an organism whose visual system has varying spatial processing fidelity (foveation) and optimizes some underlying perceptual task. If so, a computational agent equipped with human-like foveated vision and sequential eye movements optimized for a visual task should naturally produce fixation patterns that mirror those of humans.

\begin{figure}[!hbt]
    \centering
    \includegraphics[width=0.7\textwidth]{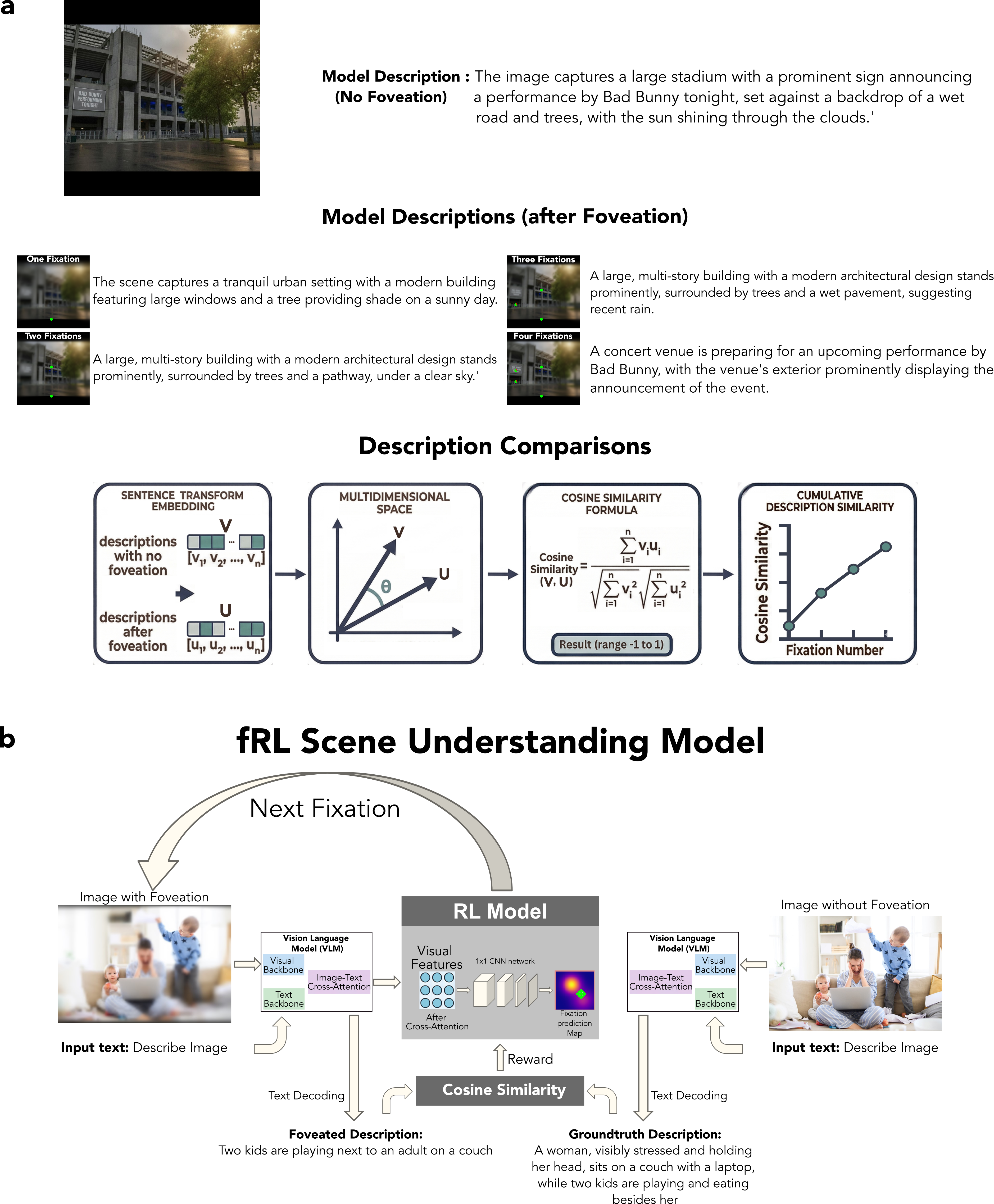}
    \caption{Semantic accuracy metric and foveated RL-scene understanding (fRL-SU) model training pipeline. (a) Schematic of a vision language model (VLM) description with and without foveation. As the number of fixations increases, the foveated description agrees more with the unfoveated description, matching it by the fourth fixation. Semantic accuracy is quantified as the cosine similarity between text-embedding vectors of the foveated and unfoveated descriptions. (b) Flowchart of the fRL-SU model's training process. After each fixation, the semantic accuracy is computed and used as a reward for the RL model. The RL model takes hidden layer features corresponding to VLM visual tokens as input, processes them using a CNN, and outputs a probability map of possible fixation locations. The RL model then selects the next fixation, and the process repeats until the model has made 4 fixations following the initial fixation.}
    \label{fig:ModelflowChart}
\end{figure}

\hfill

There are various challenges in implementing such a computational agent. The first challenge is determining what task humans might be engaged in during free-viewing. One view is that different individuals are engaged in different subtasks based on their individual-specific interests or idiosyncratic tasks. A car lover might look at a '60s classic Citroen 12V, a bird-watcher on a purple martin, and a music lover on a vintage poster of Sonic Youth on the wall. Thus, we may be unable to define a common task that governs how most people view scenes naturally. Another view suggests that a default task most humans engage in involves comprehending scenes and events \cite{loschky2020scene, coco2012scan, coco2014classification, loschky2026role}. Furthermore, a recent study showed that the locations humans look in scenes when free-viewing are similar to those where they look when explicitly instructed to describe the scene. \cite{murlidaran2025eye}. 

\hfill

Thus, a good starting point is to assess which fixation patterns are optimal for a foveated agent maximizing scene comprehension, and whether those patterns resemble human free-viewing eye movements. This leads to the second challenge in creating such a computational agent: a model that can accurately comprehend scenes and capture how different fixations (and foveations) translate into scene comprehension accuracy. Recent advances in compact visual language models (VLMs), which can describe scenes at human-level and run on local GPUs, provide a viable path to implement the model \cite{lu2024ovis, yao2024minicpm, bai2025qwen3, clark2026molmo2}. Implementing existing fixation-dependent visual loss techniques \cite{perry2002gaze} on the image can degrade the model’s scene comprehension accuracy with varying fixation points in the scene (Fig. \ref{fig:ModelflowChart}a).   

\hfill

The third challenge is how to optimize the agent’s eye movement exploration to maximize the understanding of scenes. This requires a metric to measure the accuracy of scene descriptions. We measure semantic accuracy by representing both the agent's scene description (after N fixations) and the ground-truth description (no foveation, full resolution) as text vectors (embeddings), then taking their cosine similarity (Fig. \ref{fig:ModelflowChart}a). This metric correlates well with human judgments \cite{murlidaran2025eye}. To maximize semantic accuracy, the model should use its current foveated view to select the next fixation that most improves scene description. Traditional approaches use a Bayesian ideal observer constrained by a foveated visual system and provide closed-form mathematical expressions to estimate accuracy-maximizing fixation locations or eye movement actions \cite{najemnik_optimal_2005, eckstein2015optimal, zhang2010evolution, ackermann2013choice}. These methods cannot be applied to real-world scenes without making a large number of assumptions because the probability density functions of the stimuli are unknown. Reinforcement learning techniques allow for optimization of actions by rewarding the association of actions and states \cite{weisswange2009can, sullivan2011modular, hoppe2019multi}. Neural networks provide a method to learn a mapping between multi-dimensional states describing features of a scene and actions \cite{mnih2015human}, and have been shown to approximate optimality of eye movements for simpler search tasks where the Bayesian ideal is computable \cite{zhou2022deep}.   

\hfill

Here, we combine a compact visual language model \cite{lu2024ovis}, simulated foveation \cite{perry2002gaze}, and reinforcement learning (RL) to develop a model that learns eye movements to maximize scene understanding (the foveated RL for scene understanding, or fRL-SU, model). We hypothesize that the signature fixation patterns humans exhibit when viewing real-world scenes emerge from a near-optimal eye movement strategy for scene comprehension. We note that previous studies have developed deep neural network models that can be trained to successfully predict human fixation patterns on scenes (e.g., DeepGaze \cite{kummerer2022deepgaze}, ScanDiff \cite{cartella2025modeling}, PathGan \cite{assens2018pathgan}, etc.). An important distinction is that these models are trained on large datasets of human fixations. They do not execute any perceptual task, do not incorporate the foveated processing of the human visual system, and do not estimate task-optimal eye movements. They therefore do not address how these fixation patterns could arise in an organism whose foveated visual system is optimized for ecologically relevant tasks. The fRL-SU model is not trained on human fixations. It is trained to generate fixations that optimize scene understanding under constraints of foveated vision. 

\hfill

We compare the fixations of the fRL-SU model to those of 50 human observers free-viewing 147 images.  Our main metric is the frequency of fixations to different categories: people, text, objects relevant to scene understanding, and objects gazed at or grasped by people in the scene. We reason that if humans are optimizing scene comprehension during free-viewing, then training the fRL model to maximize accuracy on other tasks, such as identifying the smallest object in the scene or categorizing the scene \cite{anderson2021category}, should predict fixations that do \textit{not} match free-viewing human patterns as well. Furthermore, if the fRL-SU model fails to predict human fixations under either minimal or extreme foveation, this would indicate that human fixation patterns are a consequence of an interaction between the human foveated visual system and scene-comprehension optimization. As additional control models, we compare the fRL-SU model to saliency models (GBVS \cite{harel2006graph} and Itti \cite{itti_model_1998}) and also DeepGaze, which serves as an upper-bound prediction model of human fixations, given that it is trained on human eye movement data. Finally, to assess the generalization of our model across observers and images, we evaluate the fRL-SU model on a recently published dataset that quantifies how fixations to text, people, and touched objects vary with age for 6,720 individuals \cite{linka2025protracted}.

\section*{Results}
\subsection*{Foveated Reinforcement Learning Scene Understanding Model (fRL-SU)}

We incorporated foveation using an image-based method that applies progressively stronger linear filtering with increasing retinal eccentricity (distance from fixation) and de-blurs the image with subsequent fixations\cite{perry2002gaze}. This might be an incomplete model of foveation, but implementing other methods, such as spatial pooling of the model's internal feature representations \cite{akbas2017object}, required extensive retraining of the VLM.  As fixations accumulate, progressively more of the image is represented at high resolution. Fig. \ref{fig:ModelflowChart}a illustrates how the VLM's description evolves as fixations accumulate. The model's degradation with retinal eccentricity was calibrated to match that of humans. This was carried out using a separate psychophysical study in which participants (N=20) viewed 277 images on a gaze-contingent display and described each image after two or four fixations. The model’s foveation was adjusted so that, when it followed the measured fixations of individual humans, the accuracy of its generated scene descriptions matched that of humans (see Materials and Methods). Once the foveation parameters were set for the separate psychophysical dataset, we trained the fRL-SU model using a CNN (four 1x1 convolutional layers with ReLU activations and a softmax output). The CNN received VLM features extracted from an image foveated at all previous fixation locations and output a probability distribution over the next fixation location that maximized scene-description accuracy. Fig. \ref{fig:ModelflowChart}b depicts a detailed flow chart of the training process. Training used 24,000 images split evenly across MS-COCO \cite{lin_microsoft_2015}, MPII Human Pose \cite{andriluka14cvpr}, and FindingEmo \cite{mertens2024findingemo}. fRL-SU was then tested on the 147 test images (not seen during RL training), generating four fixations following an initial fixation at the bottom center, with ten descriptions generated after each fixation reflecting the cumulatively deblurred image up to that point (see Materials and Methods). People were present in 85 of those images, and the remaining 62 had objects and no people. The images depicted human actions, social situations, or implied actions based on the placement of objects within the scene. To dissociate low-level saliency from scene-relevant content, we selected images that contained salient regions distant from the image center and from objects or regions critical for scene understanding (see Materials and Methods). To ensure reproducibility, we trained five independent instances of the fRL-SU model. 

\begin{figure}[!hbt]
    \centering
    \includegraphics[width=0.5\textwidth]{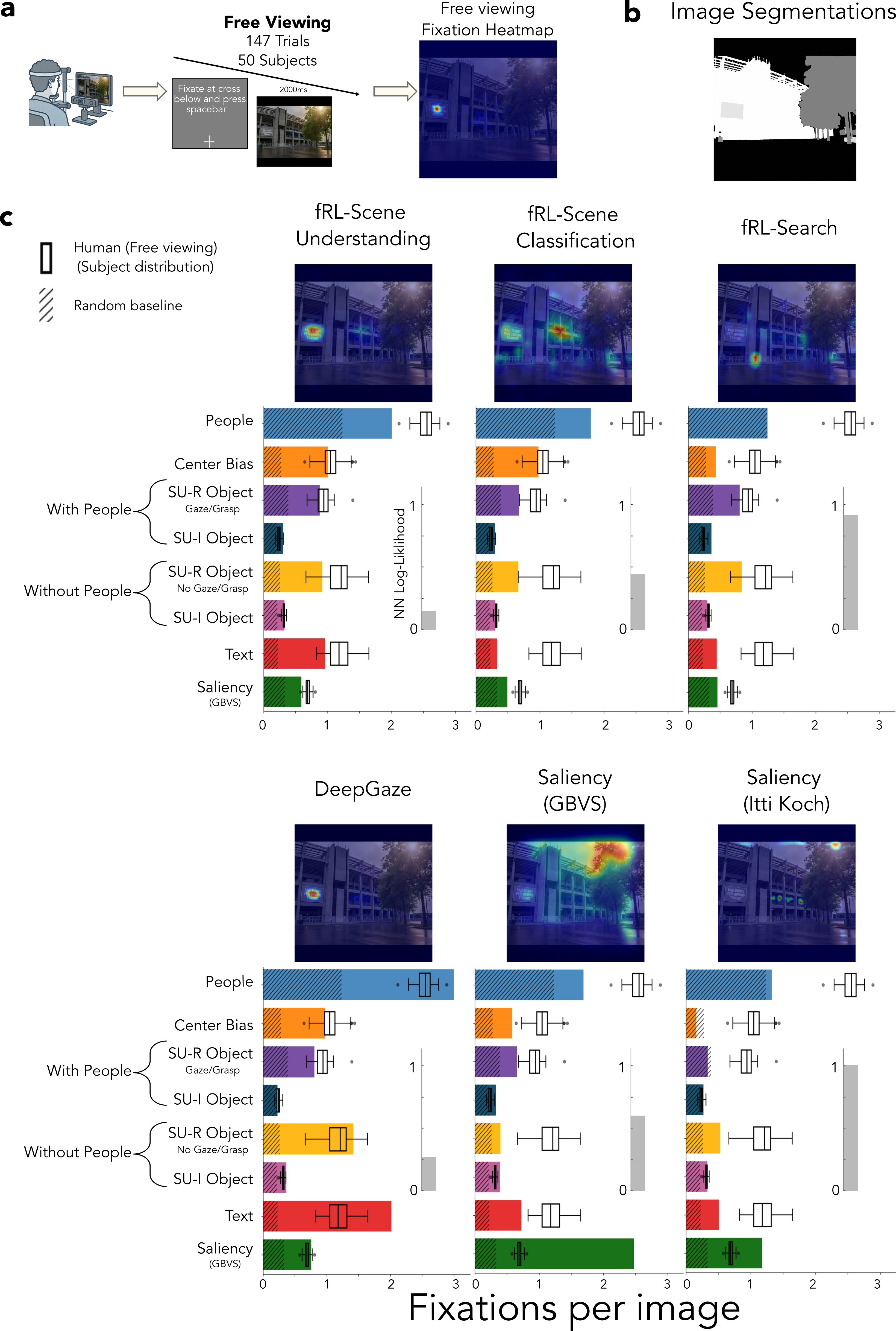}
    \caption{Fixation frequencies across scene-element categories. (a) Experimental flow for the free-viewing eye-tracking study. Fifty participants freely viewed 147 images for 2 s. (b) Example image segmentation (using SAM3) used to define scene-element categories for quantifying fixation frequencies. (c) Fixation frequencies across scene-element categories for participants (box plots) and models (bar plots). Hatched bars show a random-fixation baseline reflecting category size in the image. Heatmaps above each panel show example model predictions on the scene. The model predictions for these categories were compared with human free-viewing fixation distributions using the Negative Normalized Log Likelihood distance (NNLL), normalized to the distance observed between the Itti-Koch saliency model and human fixation distributions. The fRL-Scene Understanding (fRL-SU) model matches human data significantly better than fRL models trained on either search or scene classification. Furthermore, the fRL-SU model significantly outperforms DeepGaze and traditional saliency models.}
    \label{fig:CategoryResults}
\end{figure}

\subsection*{Human and model comparison of fixation frequencies} 
We compared the fRL-SU fixations to those of 50 participants (different from those in the foveation calibration experiment) who viewed the 147 scenes for 2 s each, with no specific instructions other than to view them freely (see Fig. \ref{fig:CategoryResults}a). We also compared the fRL-SU model to saliency models (GBVS and Itti-Koch) and with fRL models trained to maximize accuracy on alternative tasks. The fRL-Scene Classification model was trained to execute eye movements that maximize accuracy in superordinate scene classification (e.g., semantic content, 3D spatial structure, and 2D image appearance \cite{anderson2021category}). The fRL-Search model was trained to execute eye movements to maximize accuracy in identifying the smallest object in the scene.  

\hfill

Regions within the images in our dataset were segmented as shown in Fig. \ref{fig:CategoryResults}b and categorized into scene elements: people, text, and objects. We quantified the frequency of fixations on these scene elements. Fixations were counted to a scene element if it fell within 0.7 deg of the segmentation edge to match the spatial sampling of the fRL model's action space (see fig. \ref{fig:S1} for generalization of results for different spatial thresholds). Based on prior studies \cite{clarke2014deriving}, center bias was defined as fixations falling within a 2.5 deg radius circle centered on the image (see fig. \ref{fig:S2} for generalization of results for different definitions of center bias). Objects were classified as scene understanding-relevant (SU-R object) or scene understanding-irrelevant (SU-I Object). SU-R Object was defined as an object that, when digitally removed from the image, maximally changed the VLM's scene description. Scene-understanding-irrelevant objects (SU-I objects) were defined as objects that, when removed from the scene, did not impact the scene description. Classification of SU-R and SU-I objects was based on automatically segmenting objects, placing masks, deleting them, and assessing the impact of the object removal on scene description accuracy using the VLM (see Materials and Methods, and validation of method in \cite{murlidaran2025semantic}). Fixations for objects were quantified separately for scenes with and without people. In scenes containing people, the SU-R objects were typically gazed at or grasped by a person in the scene. SU-R objects in scenes without people did not include gaze or grasp cues.

\hfill

Fig. \ref{fig:CategoryResults}c shows example fixation heatmaps of each model on the sample image from Fig. \ref{fig:ModelflowChart}a. The heatmap shows how the fRL-SU and DeepGaze models fixate on the text sign on the left, matching human participants' fixations. In contrast, the other models do so to a lesser extent (RL-SC and GBVS) or not at all (RL-Search and Itti-Koch saliency). Additional fixation heatmaps for each scene-element category across all models are shown in fig. \ref{fig:S3}. Our results showed that, averaged across images, the human fixations exhibited frequency patterns similar to those observed in prior studies. People were significantly fixated more often than any other category (p $<$ 0.0001 for all comparisons). Objects critical to scene understanding were fixated more often than both unimportant objects and salient regions defined by GBVS (p$<$0.0001, for both comparisons). Regions with text were fixated more often than salient regions (p $<$ 0.0001, all the comparisons above have FDR-adjusted p-values). The fRL-SU fixation frequencies closely matched those of human participants across all scene-element categories (Fig. \ref{fig:CategoryResults}c). The model primarily fixated on people, prioritizing objects they are looking at or holding over those they are not interacting with. In scenes without people, grasp or gaze cues, the fRL-SU model still fixated more on SU-R objects than SU-I objects. The model also captured the large proportion of text fixations. In addition, first fixations were biased towards the image center, exceeding the frequency expected under the random-fixation baseline. The largest human-fRL-SU discrepancy was the model’s lower frequency of fixations on people relative to participants (see Discussion). In contrast to the fRL-SU, the saliency models generated fixation frequencies that differed from the participants' fixation frequencies. As expected, the saliency models primarily fixated on low-level salient regions and underpredicted fixation frequencies across most categories, except for SU-I objects. DeepGaze overpredicted human fixation frequencies on people, text, and SU-R objects.

\hfill

Fig. \ref{fig:CategoryResults}c also shows that the fRL-Search and fRL-SC models (control models) generated patterns closer to those of humans than the saliency models, but to a lesser degree than the fRL-SU model.  The control RL models showed reduced fixations on text, SU-R objects (those not gazed at or grasped), and people. To quantify the distance between the models and human fixation frequencies (excluding those to the most salient objects to avoid over-penalizing GBVS), we calculated a negative normalized log-likelihood (NNLL). A lower number reflects better agreement between model and human data. The fRL-SU were closer to participant fixation frequencies than the fRL-SC, fRL-Search ($p < 0.0001$ vs. fRL-SC; $p < 0.0001$ vs. fRL-search), saliency models ($p < 0.0001$ vs. GBVS; $p < 0.0001$ vs. Itti-Koch), and DeepGaze ($p < 0.0001$) due to the higher fixation frequencies than participants to categories like people, SU-R objects, and text.  The main analysis assumed that the categories were independent while computing the NNLL scores. The NNLL scores were similar even when dependencies between categories were taken into account (fig. \ref{fig:S4})

\hfill

We complemented our fixation frequency analyses by using other metrics, such as AUROC and heatmap correlations. In both AUROC and heatmap correlations, DeepGaze performed best, followed by fRL-SU ($p < 0.001$). fRL-SU significantly outperformed fRL-SC, fRL-Search, and the saliency models on both metrics ($p < 0.001$ for all comparisons, table \ref{tab:model_performance}).

\begin{figure}[!hbt]
    \centering
    \includegraphics[width=0.6\textwidth]{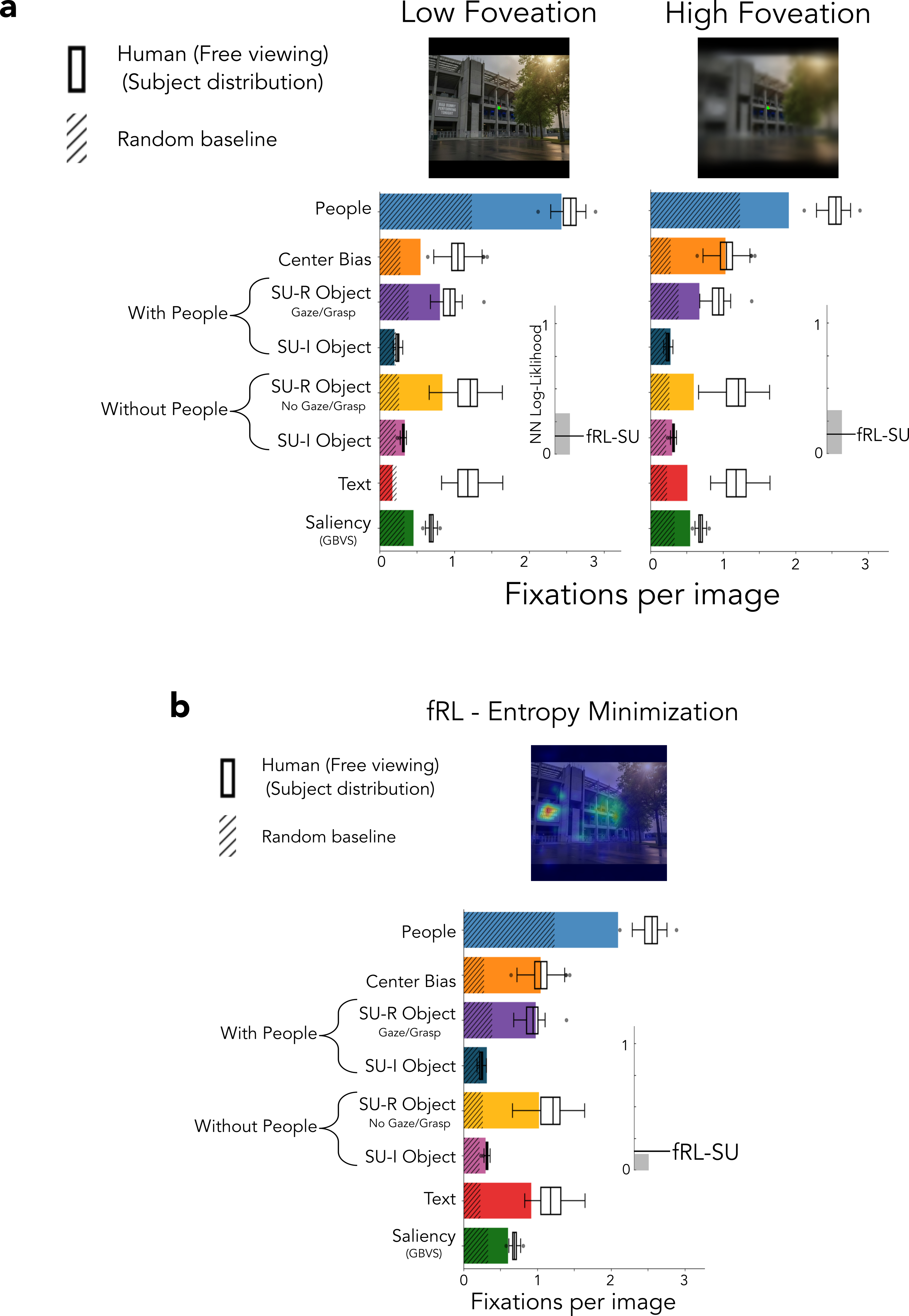}
    \caption{Effect of foveation and fRL-Entropy model results. (a) Fixation frequencies from fRL-SU models trained with extremely low (left) or high (right) foveation. Images at the top show the foveated input to the VLM at each foveation level. Both extremes yield significantly worse agreement with human fixation frequencies than the human-calibrated fRL-SU model. (b) fixation frequencies from an alternative model, fRL-Entropy, trained to minimize entropy of the VLM's next token distribution (heatmap shows example fixation density). fRL-Entropy significantly outperforms the fRL-SU model in matching human fixation frequencies.}
    \label{fig:FOV_and_Ent_Results}
\end{figure}

\subsection*{The Influence of Foveation in fRL-SU Fixation Predictions}
We hypothesized that the close correspondence between fRL-SU and human fixation frequencies is an emergent property arising from the interaction between two factors: optimization of scene understanding and the foveated nature of the human visual system. Under this hypothesis, agreement with human fixation frequencies should be lower if the fRL-SU model is trained with foveation substantially weaker or stronger than that of humans. To test this prediction, we retrained the model with extremely low and high foveation. Fig. \ref{fig:FOV_and_Ent_Results}a shows that an fRL-SU model trained with much weaker or much stronger foveation than that of humans achieved lower agreement with human fixation frequencies (p $ < $ 0.0001 for comparisons against lower and higher foveation). In particular, fixations on text and objects were reduced compared to humans, suggesting that constraints induced by foveation play a crucial role in shaping the fixation patterns observed in participants.

\subsection*{No Groundtruth Training: Foveated Reinforcement Learning with Entropy Minimization}
Our approach relied on groundtruth scene descriptions (no foveation) to train the fRL-SU model. Such a training scheme might parallel some real-world situations: a parent describing to their child what is happening in a picture book or a real-world scene provides a verbal ground truth before the child explores it with their eyes. However, access to groundtruth descriptions during training is unlikely to be the norm when children learn to explore scenes. This raises a question: can an agent with foveated vision learn to exhibit human-like eye movements without access to groundtruth descriptions? We modified the fRL-SU loss function so that the model was instead trained to minimize the VLM's uncertainty in generating descriptions. Specifically, for each foveated image, we sampled 10 high-probability yet semantically distinct candidate descriptions from the VLM and minimized the average per-token (word chunks) entropy across these candidates. (see Materials and Methods and fig. \ref{fig:S5}). Thus, the model (fRL-Entropy) learned to make eye movements to minimize this uncertainty without access to the ground truth (unfoveated descriptions). Fig. \ref{fig:FOV_and_Ent_Results}b shows the predicted fixation heatmap of the fRL-Entropy model for a sample image. The fRL-Entropy matched human fixation frequencies as well as, or even better than, fRL-SU ($p = 0.001$).   

\subsection*{Generalization to other images and eye movement datasets}
\begin{figure}[!hbt]
    \centering
    \includegraphics[width=0.75\textwidth]{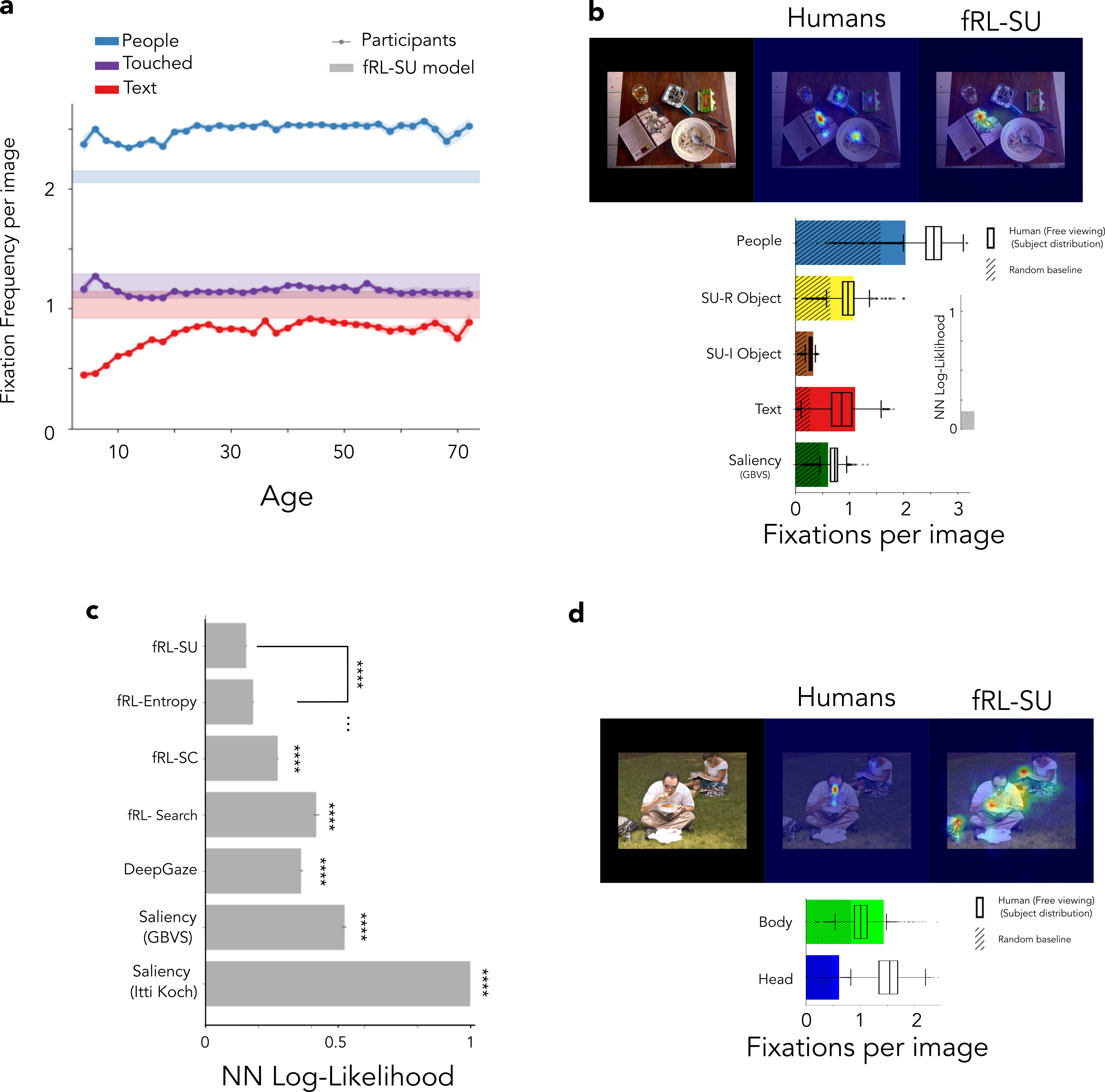}
    \caption{fRL-SU model results on a publicly available free-viewing dataset \cite{linka2025protracted} (a) The dataset contains human free-viewing data across different ages, 5-72 years (N=6,720). Solid lines show human fixation frequencies across ages for three image categories: people, touched objects, and text. Dashed lines indicate fRL-SU model predictions. (b) Applying our category definitions to this dataset yields similar fixation patterns. The fRL-SU model successfully matches human data across these categories (measured and predicted heatmaps shown). (c) The NN-Loglikelihood score shows that the fRL-SU model best matches human data, consistent with the results observed in our dataset (d). However, the model exhibits mismatched predictions regarding fixations on specific body parts. While the model correctly prioritizes people present in the scene, it focuses disproportionately on the body, whereas humans typically fixate on the head (example heatmaps shown).}
    \label{fig:DeHaas}
\end{figure}

To assess the generalization of our model and findings to other images and eye movement datasets, we evaluated the fRL-SU model's fixation frequencies on a recently published dataset \cite{linka2025protracted} of free-viewing fixations from 6,720 human observers aged 5 to 72, and on a set of 40 images intended to study how age influences free-viewing fixation (see Materials and Methods for details on this dataset). The fRL-SU model was applied with its trained parameters unchanged. We only modified the subtended angles of the images to match the range (the upper and lower bounds) of stimulus dimensions and viewing distances reported in the original study \cite{linka2025protracted}. We first compared model and human fixation frequencies. Using the three scene-element categories reported in the original study (people, grasped (touched) objects, and text), fRL-SU reproduced the relative fixation frequencies observed in human participants (Fig. \ref{fig:DeHaas}a). We then analyzed fixation frequencies, including additional categories: SU-R objects, SU-I objects (using our automated method), and the most salient region (based on GBVS, Fig. \ref{fig:DeHaas}b). The agreement between fRL-SU and human fixation frequencies of the adult population (ages 18 to 72) on this dataset was similar to that observed in our own data and significantly better than that achieved by control models (p $ < $ 0.0001 for all comparisons, Fig. \ref{fig:DeHaas}c). See fig. \ref{fig:S6} for the corresponding frequency distributions.

\section*{Discussion}

Why humans look at scenes the way they do during free-viewing and what they aim to accomplish remain poorly understood. Here, we addressed these questions by developing a computational agent with simulated human foveated vision that learns optimal (accuracy-maximizing) eye movements for different visual tasks on real-world scenes. The model optimized for scene understanding produced emergent fixation patterns that matched those of humans: increased fixation frequencies on people, text, scene-relevant objects (relative to scene-irrelevant objects), gazed or grasped objects, and even the initial center bias. This pattern held in both our own data (50 participants viewing 147 images) and a large, previously published dataset (6720 participants and 40 images). Training the model to optimize different tasks (e.g., searching for the smallest object or superordinate-level scene classification) resulted in worse predictions of human fixation frequencies. Training fRL-SU with very low or very high foveation also decreased agreement with humans, specifically reducing the model's fixations on text and scene-relevant objects. Notably, reducing the model's foveation also weakened the initial center bias, suggesting that this prevalent human behavior is not solely a byproduct of photographer bias (placing content at image centers) \cite{tatler2007central, tseng2009quantifying} but also reflects a near-optimal strategy given foveated vision: an initial central fixation minimizes the distance of all image regions to central vision. Together, these results suggest that human free-viewing fixations emerge from the interaction between optimization for scene understanding and the specific properties of the human foveated visual system.   

\hfill

The fRL-SU model outperformed DeepGaze in predicting human fixation frequencies. This finding might seem puzzling, since DeepGaze is trained on human fixations and would, by design, be expected to provide the best predictions. One explanation is that DeepGaze is trained on fixation maps accumulated over 3 to 5 seconds of viewing (about 10 to 12 fixations), whereas our analyses focus on only the first four fixations. Early human fixations are more strongly constrained by peripheral visibility and foveation than later fixations, which may explain why a model built on foveation principles (fRL-SU) better predicts early fixations than a model trained on later- or longer-time-course data (DeepGaze). This seems qualitatively consistent with the finding that DeepGaze fixates on text more frequently than humans and the fRL-SU in their first four fixations. Furthermore, when evaluating the precise location of the human fixations using AUROC and heat maps, DeepGaze outperformed the fRL-SU model, as expected. 

\hfill

An important finding is that an fRL model that learns eye movements with a reward function that minimizes sentence entropy also resulted in emergent human-like fixations. This might provide a way for humans to learn optimal fixations for scene comprehension without a priori access to the ground truth before eye-movement exploration.

\hfill

The fRL-SU or fRL-Entropy results should not be interpreted as suggesting that language is needed to learn eye movements that optimize scene comprehension. That would imply that humans who never acquire language would be unable to learn to optimize eye movements for scene comprehension, which is unlikely. Language is a convenient representation of scene meaning for training, but visual representations aligned with language's semantic space should yield similar emergent fixation patterns (except for text, which inherently requires a language representation).

\hfill

One limitation of our modeling effort is that the fRL-SU model resulted in fewer fixations on people than human observers for both our dataset and that of Linka et al. \cite{linka2025protracted}. Furthermore, a breakdown of fixations to bodies and heads shows a dissociation. The fRL-SU model fixated on bodies more often, while humans fixated on faces more frequently (see Fig. \ref{fig:DeHaas}d). This suggests that humans might not be solely optimizing scene comprehension when free-viewing. There might be a complementary task to identify people, their gender, age, and affective state. When the fRL model was implemented with two tasks: scene understanding and identifying a person's age, emotion, and gender, it fixated on people more frequently, yielding results comparable to human gaze patterns (see fig. \ref{fig:S7}).  However, even that model did not attain the high fixation frequency on faces that humans showed. Two factors contribute to this limitation. One, the VLMs we used are compact, can be implemented in our local GPU cluster to speed up the RL training, but do not reach human understanding of the real world \cite{rosenberg2026limits, wichmann2023deep}, including not achieving human-level face-processing performance. Second, our model relies on a simplified linear filtering approach for foveation. As a result, it does not account for the non-linear effects of crowding, which selectively degrade the peripheral processing of text and faces  \cite{martelli2005faces} and might influence the learned eye movement strategy. 

\hfill

In conclusion, our results indicate that eye movements during free-viewing of scenes reflect a task-driven strategy shaped by the goal of comprehending the visual world. The characteristic pattern of fixations on people, text, scene-relevant objects, and gazed or grasped objects emerges from the interaction between scene-understanding optimization and the foveated architecture of the human visual system.


\clearpage 

%
\bibliography{science_template} 
\bibliographystyle{sciencemag}

%
%
%
%
%
%


\section*{Acknowledgments}
We would like to thank Professor William Wang of the Computer Science Department for his invaluable support and guidance throughout this project. We also extend our gratitude to our lab members, Srijita, Parsa, and Anqi, for their thoughtful insights and suggestions throughout this work.
\paragraph*{Funding:}
This research received no specific grant from any funding agency in the public, commercial, or not-for-profit sectors 
\paragraph*{Author contributions:}
S.M. and M.P.E. contributed equally to the study design, experimentation, data collection, analysis, and manuscript writing.
Z.W. provided the AutoSUM measurements for the test dataset and assisted with proofreading the Methods section.
S.S. generated images for the test dataset and proofread the Introduction, Results, Figure captions, and Discussion sections.

\paragraph*{Competing interests:}
There are no competing interests to declare.
\paragraph*{Data and materials availability:}
Data, Images, model weights, and code to run the models will be publicly available upon acceptance.


\subsection*{Supplementary materials}
Materials and Methods\\
Table S1\\
Figs. S1 to S7\\


\newpage


\renewcommand{\thefigure}{S\arabic{figure}}
\renewcommand{\thetable}{S\arabic{table}}
\renewcommand{\theequation}{S\arabic{equation}}
\renewcommand{\thepage}{S\arabic{page}}
\setcounter{figure}{0}
\setcounter{table}{0}
\setcounter{equation}{0}
\setcounter{page}{1} 


\begin{center}
\section*{Supplementary Materials for\\ \scititle}

Shravan~Murlidaran$^{\ast\dagger}$,
Ziqi~Wen,
Sana~Shehabi,
Miguel~P.~Eckstein$^{\dagger}$\\
\small$^\ast$Corresponding author. Email: smurlidaran@ucsb.edu\\
\small$^\dagger$These authors contributed equally to this work.
\end{center}

\subsubsection*{This PDF file includes:}
Materials and Methods\\
Table S1\\
Figures S1 to S7\\


\newpage


\section*{Materials and Methods}

\subsection*{Dataset}
\subsubsection*{Training Set}
To train a model to make eye movements while understanding scenes, we required a dataset of images depicting complex social activities, as such scenes would require the model to explore different regions of the image to understand what is happening. To this end, we combined images from three datasets: the MPII Human Pose dataset \cite{andriluka14cvpr}, developed by the Max Planck Institute for Informatics (MPII), which comprises 25,000 images featuring over 40,000 people performing 410 different activities; the FindingEmo dataset \cite{mertens2024findingemo}, consisting of 25,000 images from online sources that depict complex emotions among people in naturalistic settings; and the MSCOCO dataset \cite{lin_microsoft_2015}, consisting of 330,000 images commonly used for image captioning. A subset of 8,000 images was selected from each dataset (see below for details on image selection), resulting in a final training set of 24,000 images.

\subsubsection*{Testing Set}
To test our model, we curated a dataset comprising 147 images: 24 from the MS-COCO validation set, 47 screenshots of movie scenes depicting complex social interactions, and 76 images that were a mix of in-house photographs and AI-generated images created with tools such as Google Nano Banana Pro. The images were chosen so that the semantic accuracy of the VLM model's descriptions relative to the human ground-truth description exceeded 0.5. We used embedding models and computed cosine similarity between description embeddings to determine the score. The images were also ensured to have at least $2-DVA$ between the most salient and critical regions of the scene.

\subsection*{Eye Movement Study}
Using the test dataset, we conducted a free-viewing study in which 50 participants viewed each scene for 2 seconds without any explicit task. We tracked eye movements using an EyeLink 1000+ desktop-mounted system (spatial resolution 0.01°) operating at 2,000 Hz. Participants were positioned 75 cm ($\bm{D_{obs}}$) from a 19-inch display, which subtended a visual angle of 26.6° × 21.8° at a resolution of 1,280 × 1,024 (pixel size ($\bm{s}$) = 0.0293 cm). The stimulus images were scaled to fit the monitor width (without altering the aspect ratio) and were centrally positioned. Participants fixated a cross located 480 pixels (10.1°) below the center of the monitor and pressed the spacebar to begin each trial. To reduce head motion, participants used a chin rest. Saccades were identified using velocity (22°/s) and acceleration (4,000°/s\textsuperscript{2}) thresholds. Recordings were taken from the left eye, and the experiment was run using SR Research Experiment Builder.

\subsection*{Modeling Foveation}
\subsubsection*{Simulation:}

\hfill

\noindent To simulate foveation, we require a method that degrades the information available in an image as a function of eccentricity from the fixation location. To this end, we employed the method developed by Perry and Geisler \cite{perry2002gaze} and later implemented by Jiang et al. \cite{jiang2015salicon, ouyang2018imagefoveation} for modeling foveation while exploring scenes through mouse movements. Their method applies Gaussian blurring to the image, with the blur bandwidth (half-max) increasing as a function of eccentricity from the fixation point. This function is referred to as the resolution map ($\bm{R}$). Jiang et al. \cite{jiang2015salicon} define the following resolution map:

\hfill

\begin{equation}
\left.
\begin{aligned}
\bm{R}(\bm{x},\bm{y}) &= \frac{\bm{\alpha}}{\bm{\alpha} + \bm{\theta}(\bm{x},\bm{y})}, \\[6pt]
\bm{\theta}(\bm{x},\bm{y}) &= \frac{\bm{D}(\bm{x},\bm{y})}{\bm{p}}, \\[10pt]
\end{aligned}
\right.
\end{equation}
\begin{align*}
\textbf{where,} & \\
\bm{x},\bm{y} &\text{ are pixel coordinates in the image,} \\
\bm{\alpha} &\text{ is a hyper-parameter controlling how quickly resolution falls off with} \\
&\text{eccentricity (smaller } \bm{\alpha} \text{ yields a faster drop),} \\
\bm{\theta}(\bm{x},\bm{y}) &\text{ is the eccentricity of pixel location } (\bm{x},\bm{y}) \text{ from the fixation point, } \\
&\text{expressed in degrees of visual angle,} \\
\bm{D}(\bm{x},\bm{y}) &\text{ is the Euclidean distance of location } (\bm{x},\bm{y}) \text{ from the fixation location,} \\
\bm{p} &\text{ is the number of pixels per degree of visual angle, and} \\
\bm{R}(\bm{x},\bm{y}) &\text{ is the resolution map (half-max bandwidth at } (\bm{x},\bm{y})\text{).}
\end{align*}
\hfill

\noindent In vision research, the number of pixels per degree ($\bm{p}$) is a critical metric used to standardize stimulus size relative to the observer's perspective. It is calculated from the observer distance ($\bm{D_{obs}}$) and the physical pixel size ($\bm{s}$) using the following geometric relationship:

\hfill

\begin{equation}
\left.
\begin{aligned}
\bm{p} &= \left\lfloor \frac{\bm{D_{obs}} \cdot \tan(1^\circ)}{\bm{s}} \right\rfloor, \\[10pt]
\textbf{where,} & \\
\bm{p} &\text{ is the number of pixels per degree of visual angle,} \\
\bm{D_{obs}} &\text{ is the observer distance (distance from eye to screen),} \\
\bm{s} &\text{ is the physical size of a single pixel, and} \\
\lfloor \dots \rfloor &\text{ denotes the floor function (integer truncation).}
\end{aligned}
\right.
\end{equation}

\hfill

\subsubsection*{Approximating Gaussian Blur Using Gaussian Pyramids:}

\hfill

\noindent Convolving an image with a Gaussian kernel to produce blur is computationally expensive, with even efficient implementations typically running in $\bm{O(N \log N)}$ time, where $\bm{N}$ is the number of pixels in the image. To address this, Perry and Geisler \cite{perry2002gaze} introduced the use of Gaussian pyramids, a multi-scale representation in which the image is iteratively downsampled by a factor of two and then upsampled back to the original resolution. Perry and Geisler empirically showed that this process approximates Gaussian blurring with specific half-max bandwidths. These functions are referred to as the transfer functions ($\bm{T_i}$). The standard deviations of these transfer functions in spatial frequency are given by:

\hfill

\begin{equation}
\left.
\begin{aligned}
\bm{\sigma}_{\bm{T}_1} &\approx 0.124 \ \text{cycles/pixel}, \\
\bm{\sigma}_{\bm{T}_2} &\approx 0.056 \ \text{cycles/pixel}, \\
\bm{\sigma}_{\bm{T}_3} &\approx 0.0267 \ \text{cycles/pixel}, \\
\bm{\sigma}_{\bm{T}_i} &\approx \frac{\bm{\sigma}}{2^{\bm{i}}} \ \text{cycles/pixel}, \\[10pt]
\textbf{where,} & \\
&\text{the first three values are empirically derived from Perry \& Geisler \cite{perry2002gaze},} \\ &\text{and the general formula approximates the pattern across pyramid levels.} \\
\bm{\sigma}_{\bm{T}_i} &\text{ denotes the standard deviation of transfer function } \bm{T}_i \\ 
&\text{ in spatial frequency,} \\
\bm{\sigma} &\text{ is the standard deviation of the base Gaussian blur, and} \\
\bm{i} &\text{ indexes the Gaussian pyramid level.}
\end{aligned}
\right.
\end{equation}

\hfill

\noindent $\bm{T_0}$ corresponds to the original image and can be assumed to be part of the Gaussian pyramid, originating from an even larger pyramid. Therefore, the standard deviation of $\bm{T_0}$ can be taken as $0.248$. By interpolating between these predefined transfer functions, one can efficiently approximate a Gaussian blur of any desired half-max bandwidth, thereby achieving linear time complexity. Following Perry and Geisler, the transfer functions are expressed on a relative spatial frequency axis, where $1.0$ corresponds to $0.25$ cycles/pixel. The function used in our study is defined as:

\hfill

\begin{equation}
\left.
\begin{aligned}
\bm{T}_{\bm{i}}(\bm{f}) &=
\begin{cases}
e^{- \left((\bm{f}/4) / (\bm{\sigma} / 2^{\,\bm{i}}) \right)^2}, & \bm{i} = 1, \ldots, 10, \\[7pt]
0, & \bm{i} = 11,
\end{cases}
\\[10pt]
\textbf{where,} & \\
\bm{T}_{\bm{i}}(\bm{f}) &\text{ is the transfer function at pyramid level } i, \\
\bm{f} &\text{ is the relative spatial frequency, normalized such that } f = 1.0 \\ 
&\text{ corresponds to } 0.25 \ \text{cycles/pixel,} \\
\bm{\sigma} &\text{ is the standard deviation of the base Gaussian blur in spatial} \\
&\text{frequency }(\sigma = 0.248 \ \text{cycles/pixel for } T_0), \text{ and} \\
\bm{i} &\text{ is the index of the Gaussian pyramid level } (i=1 \text{ is the finest blur,} \\
&\text{ and larger }i \text{ correspond to coarser scales.})
\end{aligned}
\right.
\end{equation}

\hfill

\subsubsection*{Interpolation Between Transfer Functions:}

\hfill

\noindent To achieve arbitrary levels of blurring, the following interpolation function is used:

\hfill

\begin{equation}
\left.
\begin{aligned}
\bm{B}_{\bm{i}}(\bm{x},\bm{y}) &=
\begin{cases}
0, & \bm{R}(\bm{x},\bm{y}) \leq \bm{R}_{\bm{i}}, \\[8pt]
\dfrac{0.5 - \bm{T}_{\bm{i}}\!\left(\bm{R}(\bm{x},\bm{y})/2\right)}{\bm{T}_{\bm{i-1}}\!\left(\bm{R}(\bm{x},\bm{y})/2\right) - \bm{T}_{\bm{i}}\!\left(\bm{R}(\bm{x},\bm{y})/2\right)}, & \bm{R}_{\bm{i}} < \bm{R}(\bm{x},\bm{y}) < \bm{R}_{\bm{i-1}}, \\[12pt]
1, & \bm{R}(\bm{x},\bm{y}) \geq \bm{R}_{\bm{i-1}},
\end{cases}
\\[10pt]
\end{aligned}
\right.
\end{equation}
\begin{align*}
\textbf{where,} & \\
\bm{B}_{\bm{i}}(\bm{x},\bm{y}) &\text{ is the interpolation weight at pixel } (x,y) \text{ for pyramid level } i, \\
\bm{R}(\bm{x},\bm{y}) &\text{ is the half-max bandwidth given by the resolution map,} \\
\bm{R}(\bm{x},\bm{y})/2 &\text{ converts this bandwidth to the normalized frequency axis used} \\ &\text{by the transfer functions} \\
\bm{R}_{\bm{i}}, \bm{R}_{\bm{i-1}} &\text{ are the half-max bandwidths corresponding to transfer} \\
&\text{functions } T_i \text{ and } T_{i-1}, \\
\bm{T}_{\bm{i}}\!\left(\tfrac{\bm{R}(\bm{x},\bm{y})}{2}\right) &\text{ is the Gaussian transfer function at scale } i \text{ evaluated at the} \\
&\text{half-max bandwidth } R(x,y), \text{ and } 0.5 \text{ is the half-maximum} \\
&\text{sensitivity used to interpolate between adjacent scales.}
\end{align*}
\subsubsection*{Multiple Fixations:}
\noindent To incorporate multiple fixations, the eccentricity maps $\bm{\theta}_{\bm{j}}(\bm{x},\bm{y})$ generated by each fixation are combined using the following equation:
\begin{equation}
\left.
\begin{aligned}
\bm{\theta}_{\text{combined}}(\bm{x},\bm{y}) &= \min\big(\bm{\theta}_{1}(\bm{x},\bm{y}),\space \bm{\theta}_{2}(\bm{x},\bm{y}), \ldots, \bm{\theta}_{\bm{N}}(\bm{x},\bm{y})\big), \\[10pt]
\textbf{where,} & \\
\bm{N} &\text{ is the number of fixations made, and} \\
\bm{\theta}_{\bm{j}}(\bm{x},\bm{y}) &\text{ is the eccentricity map for the } j^{\text{th}} \text{ fixation.}
\end{aligned}
\right.
\end{equation}
\noindent This combined eccentricity map effectively deblurs all previously fixated locations. This provides a convenient way of combining the information gained from multiple fixation locations into a single foveated representation.

\subsection*{Vision Language Model (VLM)}
In this study, we use Ovis2.5 \cite{lu2024ovis, lu2025ovis25technicalreport} as our base vision-language model (VLM). Ovis2.5 is a decoder-only transformer that accepts multimodal inputs (text, images, and videos) and produces multimodal outputs in the same domains. Here, we restrict the inputs to images (both foveated and unfoveated) and text, and generate two complementary outputs: natural-language descriptions of the scene (the standard capability of the model) and predictions of the next fixation location (our novel extension).

\hfill

\noindent Internally, Ovis2.5 incorporates the Native Vision Transformer (NaViT) \cite{dehghani2023patch} for image encoding. NaViT partitions an input image into contiguous $16 \times 16$ pixel patches, transforms each patch into a feature vector, and then projects these vectors into a 4096-dimensional embedding space shared across modalities. For text, Ovis2.5 relies on the Qwen3 large language model \cite{yang2025qwen3}, which likewise projects token embeddings into a 4096-dimensional space. The vision and language embeddings are concatenated and passed into a cross-attentional decoder. In Ovis2.5, the decoder produces a corresponding 4096-dimensional output vector for every input patch or text token embedding. The outputs aligned with image tokens are routed into our eye movement model to predict the next fixation, while the decoder outputs at text positions are passed through the language modeling head to generate the sentence.

\subsection*{Eye Movement Model}
The eye movement model uses average spatial pooling with a stride length of two on the input image tokens from Ovis2.5, followed by a Convolutional Neural Network (CNN) with Pointwise Convolution, i.e., convolution layer with $1 \times 1$ kernel, combined with layer normalization and ReLU activations, to sequentially project the 4096-dimensional decoder feature vector for each image token into a lower-dimensional space:




\hfill

\begin{equation}
\left.
\begin{aligned}
\mathbf{Input} &\in \mathbb{R}^{4096 \times 2H \times 2W} \xmapsto{\text{Spatial Pooling}} \mathbb{R}^{4096 \times H \times W} \\[6pt]
&\qquad \mapsto \mathbb{R}^{1024 \times H \times W} \mapsto \mathbb{R}^{256 \times H \times W} \mapsto \mathbb{R}^{64 \times H \times W} \mapsto \mathbb{R}^{1 \times H \times W} \\[10pt]
\textbf{where,} & \\ \quad H  &\text{ and } W \text{ are the height and width of the action space.} \\
\end{aligned}
\right.
\end{equation}

\hfill

\noindent Each decoder feature corresponding to an image patch encodes not only the visual content of that patch but also its semantic and spatial relationships with other patches. The initial average pooling halves the number of input image tokens, thereby reducing the total computational cost of the subsequent convolution layers. To predict eye movements, it is crucial to preserve the spatial alignment while extracting the relevant features. The Pointwise Convolutions ensure that information is processed independently for each token (i.e., without pooling across neighboring patches), while the dimensionality reduction progressively distills the features into a single scalar value per patch. These scalars form a map over the pooled image tokens ($\bm{M \in \mathbb{R}^{H \times W}}$), which is then passed through a softmax layer to produce a probability distribution over possible fixation locations $\bm{p(x, y)}$. The model outputs the next fixation location ($\bm{X, Y}$) by either sampling (during training) or greedy selection (during inference) (i.e., $\bm{(X, Y) = \arg\max_{x,y} p(x, y)}$) from this probability distribution.

\hfill


\begin{equation}
\left.
\begin{aligned}
\bm{p}(\bm{x}, \bm{y}) &= \frac{\exp(\bm{M}_{\bm{x}, \bm{y}} / \bm{\tau})}{\sum_{\bm{i}=1}^{\bm{H}} \sum_{\bm{j}=1}^{\bm{W}} \exp(\bm{M}_{\bm{i}, \bm{j}} / \bm{\tau})}  \\[10pt]
\textbf{where,} & \\ 
\bm{\tau} &\text{ is the temperature parameter controlling } \\
&\text{the sharpness of the softmax distribution.}
\end{aligned}
\right.
\end{equation}


\hfill

To determine the size of the action space, we note that the VLM maps the input image from pixel space to token space with a reduction factor of 32, and the subsequent pooling introduces an additional reduction factor of 2. Thus, for an image of height $\bm{H_{image}}$ and width $\bm{W_{image}}$, the action space is given by:

\hfill

\begin{equation}
\left.
\begin{aligned}
&\text{action space} = \frac{\bm{H_{image}} \times \bm{W_{image}}}{64^2}, \\[6pt]
\textbf{where,} & \\ \text{ the factor } &64 \text{ arises from a } 32\times \text{ tokenization reduction followed by } \\ 
&2\times \text{ spatial pooling.}
\end{aligned}
\right.
\end{equation}

\hfill

\noindent To standardize the action space across images of different sizes, we define a fixed field of view in degrees of visual angle (DVA) within which all images are embedded (see below).

\hfill

\noindent Finally, to simplify training, we train a separate model for each fixation step, all sharing the same architecture described above. Thus, for $\bm{N}$ fixations, we train $\bm{N}$ independent models.

\subsection*{Reinforcement Policy}

To train our model, we need to assign rewards to the actions taken by the model. The goal of our eye movement model is to make eye movements that maximize scene understanding. To facilitate this, we convert the VLM into a scene-understanding model by providing the following text prompts along with the image it has to interpret:

\begin{center}
\textbf{Text prompt 1:} \textit{"Provide your best guess of what is happening in the scene in a sentence. Do not mention the blur seen in the picture."}
\end{center}

\begin{center}
\textbf{Text prompt 2:} \textit{"In one sentence, describe what is happening in the scene and, if people are present, identify the person if possible, determine how they appear to feel, and their gender. Do not mention the blur seen in the picture. If you do not know who the person is, simply refer to them as man, woman, girl, boy, or person."}
\end{center}

\begin{figure}[htbp]
    \centering
     \includegraphics[width=0.7\textwidth]{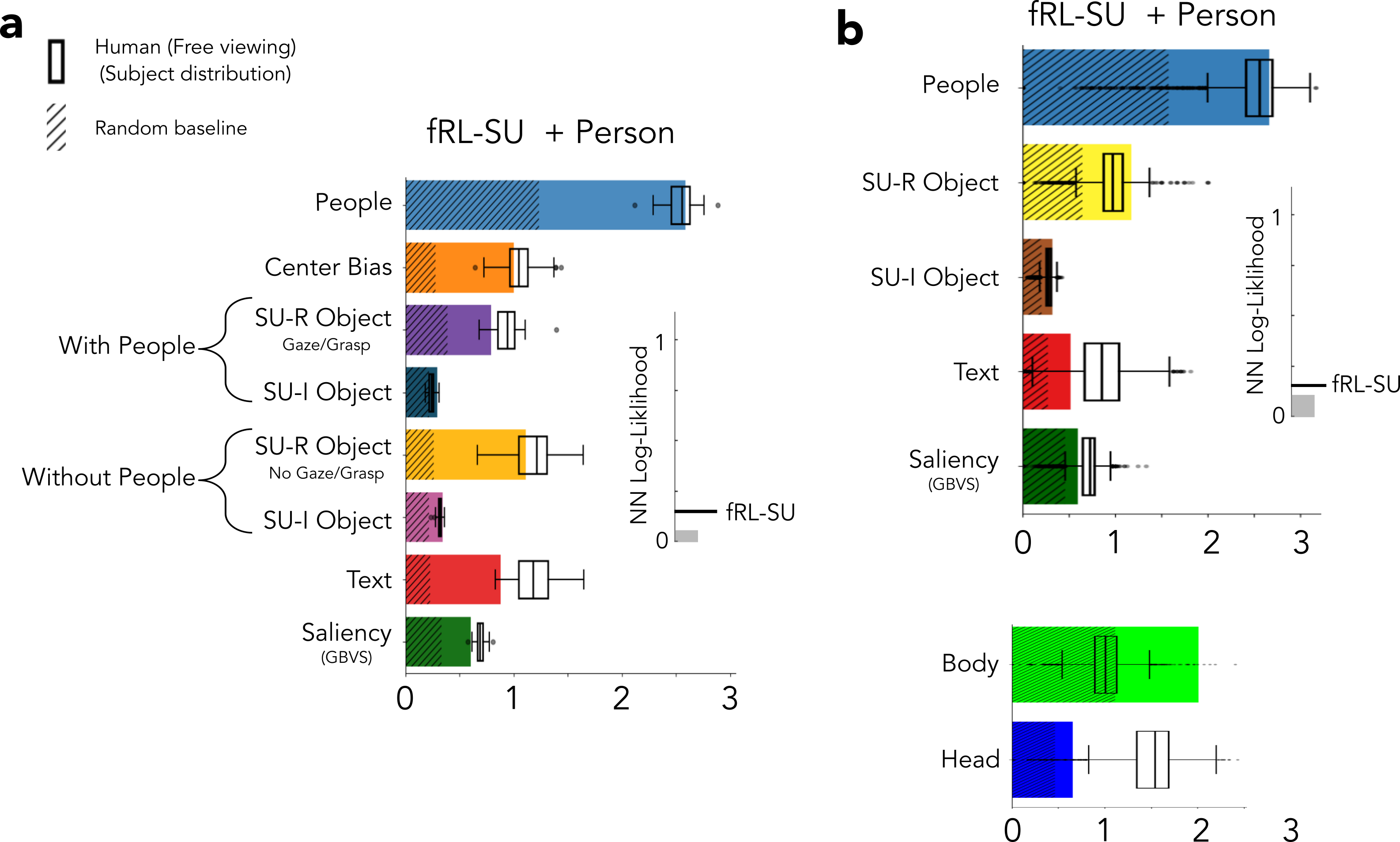} 
    \caption{Model results for detailed prompt. The fRL-SU model trained on a prompt asking the model to provide the age, gender, and emotional expressions of people while describing the scene increased the fixations to people in both (a) our dataset and (b) the OSIE short dataset from \cite{linka2025protracted}. However, the model still looks at bodies disproportionately more than faces.}
    \label{fig:S7}
\end{figure}

\noindent The main analysis in this paper covers the results for prompt 1. The results for prompt 2 are provided in fig. \ref{fig:S7}. The prompt includes an instruction not to mention blur because VLMs often explicitly describe visual artifacts such as blur when they are present in an image. Since our foveation method introduces blur in the peripheral regions of the image, we include this instruction to ensure that the model focuses on describing the scene content rather than commenting on the blur introduced by foveation.

\hfill

\noindent Now suppose the model has made $\bm{j}$ fixations ($\bm{j \leq N}$), where the initial fixation corresponds to $\bm{j=0}$. Using the foveation technique discussed above, this corresponds to deblurring $\bm{j+1}$ regions on the image (at $\bm{j}$ model fixations and one initial fixation). With this setup, an ideal reward function should satisfy the following properties:

\begin{itemize}
\item Fixating regions whose information content is important for understanding the scene should yield a high reward. Since we want the model to preferentially fixate such informative regions, the reward function should assign higher values to fixations that reveal useful scene information.

\item The cumulative reward should be monotonically increasing as a function of the number of fixations. In other words, the reward after a single fixation should be the lowest, and subsequent fixations should increase or at least maintain the reward. This reflects the intuition that revealing additional regions of the image through deblurring should not decrease the information available about the scene.
\end{itemize}

\noindent Using the VLM's descriptions after $\bm{j+1}$ fixations, we explored the following options as possible reward functions that satisfy the above conditions.

\subsubsection*{Semantic Accuracy of Descriptions}

We get semantic accuracy by computing similarity between the VLM's descriptions when the image is foveated (an image with $\bm{j+1}$ deblurred regions), and the descriptions generated when the image is unfoveated (the intact original image), which represents the maximum level of understanding the model can achieve for that image.

\hfill

\noindent For each image, we generate $\bm{M}$ descriptions for both the foveated and unfoveated conditions after each fixation. These descriptions are converted into embeddings using a Large Language Model called Jasper \cite{zhang2024jasper}. Jasper is an open-source model trained to produce sentence embeddings by distilling representations from larger closed-source embedding models. The embeddings are then used to compute the mean pairwise cosine similarity ($\bm{CS}$) between the foveated and unfoveated descriptions.

\hfill

\noindent Let $\bm{E_i}$ and $\bm{E^{fov}_{ij}}$ denote the embeddings of the $\bm{i^{\text{th}}}$ unfoveated description and the $\bm{i^{\text{th}}}$ foveated description after $\bm{j}$ model fixations. Then the cosine similarity between the two sets of descriptions is given by:

\hfill

\begin{equation}
\left.
\begin{aligned}
\bm{CS}(\bm{j}) &= \frac{1}{\bm{M}}\sum_{\bm{i}=1}^{\bm{M}}\frac{\bm{E}^{fov}_{\bm{i}\bm{j}} \cdot \bm{E}_{\bm{i}}}{\|\bm{E}^{fov}_{\bm{i}\bm{j}}\|\|\bm{E}_{\bm{i}}\|}, \\[10pt]
\textbf{where,} & \\
\bm{E}_{\bm{i}} &\text{ is the embedding of the } \bm{i^{\text{th}}} \text{ unfoveated description,} \\
\bm{E}^{fov}_{\bm{i}\bm{j}} &\text{ is the embedding of the } \bm{i^{\text{th}}} \text{ foveated description after } \\ &\bm{j} \text{ model fixations, and} \\
\bm{M} &\text{ is the number of generated descriptions.}
\end{aligned}
\right.
\end{equation}

\hfill

\noindent The cosine similarity after $\bm{j}$ model fixations, defined above, acts as a cumulative metric indicating how effective the first $\bm{j}$ model fixations are in revealing information needed to understand the scene. To compute the reward $\bm{R(j)}$ for choosing model fixation $\bm{j}$, we measure how much the current cosine similarity improves upon the best similarity obtained from previous fixations. In other words, a positive reward is obtained only when the current fixation reveals additional information that increases the similarity between the foveated descriptions and the unfoveated descriptions beyond what has already been achieved by earlier fixations.

\hfill

\begin{equation}
\left.
\begin{aligned}
\bm{R}(\bm{j}) &= \bm{CS}(\bm{j}) - \max_{\bm{k}<\bm{j}}\big(\bm{CS}(\bm{k})\big), \qquad \bm{j} \geq 1, \\[6pt]
\bm{R}(0) &= \bm{CS}(0), \qquad \text{initial fixation} \\[10pt]
\textbf{where,} & \\
\bm{CS}(\bm{j}) &\text{ denotes the cosine similarity after } \bm{j} \text{  model fixations.} & \\
\bm{R}(\bm{j}) &\text{ denotes the semantic reward for choosing } \bm{j^{th}} \text{  model fixation.}
\end{aligned}
\right.
\end{equation}

\hfill

\noindent Theoretically, as we increase the number of fixated locations, the model descriptions can ultimately approach the unfoveated descriptions (i.e., the case in which the entire image is deblurred), thereby representing the upper limit of scene understanding. Therefore, to standardize the computed rewards across images, we normalize the rewards using the mean pairwise cosine similarity among the unfoveated descriptions, denoted by $\bm{CS_{\text{upper limit}}}$. To ensure boundedness, the normalized rewards are clipped to lie between 0 and 1:

\hfill

\begin{equation}
\left.
\begin{aligned}
\bm{R}_{\text{norm}}(\bm{j}) &= \frac{\bm{R}(\bm{j})}{(\bm{CS}_{\text{upper limit}} - \bm{R}(0)) + \epsilon}, \qquad \bm{j} \geq 1, \\[6pt]
\bm{R}_{\text{norm}}(0) &= 0, \qquad \text{initial fixation}\\[6pt]
\bm{R}_{\text{norm}}(\bm{j}) &= 1, \qquad \forall \ \bm{R}_{\text{norm}}(\bm{j}) > 1, \\[6pt]
\bm{R}_{\text{norm}}(\bm{j}) &= 0, \qquad \forall \ \bm{R}_{\text{norm}}(\bm{j}) < 0, \\[10pt]
\textbf{where,} & \\
\bm{CS}_{\text{upper limit}} &\text{ is the mean pairwise cosine similarity computed} \\ 
&\text{among the unfoveated descriptions for a given image.} & \\
\bm{R_{\text{norm}}(j)} &\text{ is the normalized semantic reward for choosing the } \bm{j^{th}}\\ &\text{model fixation.} \\
\bm{\epsilon} &\text{ is a small constant for numerical stability}
\end{aligned}
\right.
\end{equation}

\hfill

\subsubsection*{Conditional Entropy}
Another possible reward function stems from the idea that the uncertainty introduced by foveation in the image pixel space should translate into uncertainty in the model's generated descriptions. Since VLMs such as Ovis2.5 are next-token predictors given an input, description generation involves predicting tokens that correspond to words. Each predicted token is associated with a probability distribution over the vocabulary, which can be used to compute the entropy rate.

\hfill

\noindent Let $\mathbf{X_0, X_1, \dots, X_T}$ denote the sequence of tokens forming a generated sentence, where $\mathbf{X_t}$ is the $\mathbf{t^{\text{th}}}$ token predicted given input image $\mathbf{I}$ and the previously predicted tokens $\mathbf{\{X_0, X_1,}$ $\mathbf{\ldots, X_{t-1}\}}$. The conditional entropy rate of the generated sequence can be written as:

\hfill

\begin{equation}
\left.
\begin{aligned}
\bm{H}(\bm{X}_0, \bm{X}_1, \ldots, \bm{X}_T \mid \bm{I}) 
&= \sum_{\bm{t}=0}^{\bm{T}} \bm{H}\!\left(\bm{X}_t \,\middle|\, \bm{X}_0, \ldots, \bm{X}_{\bm{t}-1}, \bm{I}\right), \\[6pt]
\bm{H}(\bm{X}_t \mid \bm{X}_0, \ldots, \bm{X}_{\bm{t}-1}, \bm{I}) 
&= - \sum_{\bm{k} \in \bm{\mathcal{V}}} \bm{p}_{\bm{k}} \log \bm{p}_{\bm{k}}, \\[10pt]
\textbf{where,} & \\
\bm{p}_{\bm{k}} &= p(x_t = k \mid X_0, \ldots, X_{t-1}, I), \\
\bm{\mathcal{V}} &\text{ is the vocabulary (i.e., the set of all possible tokens), and} \\
\bm{H}(\cdot) &\text{ denotes Shannon entropy.}
\end{aligned}
\right.
\end{equation}

\hfill

\noindent For $\bm{M}$ generated sentences from the VLM after $\bm{j}$ model fixations, the average entropy rate $\bm{\bar{H}(j)}$ is given by:

\hfill

\begin{equation}
\left.
\begin{aligned}
\bar{\bm{H}}(\bm{j})
&= \frac{1}{\bm{M}} \sum_{\bm{m}=1}^{\bm{M}} \frac{1}{\bm{T}_{\bm{m}\bm{j}}} \bm{H}\!\left(\bm{X}^{(\bm{m}\bm{j})}_0, \bm{X}^{(\bm{m}\bm{j})}_1, \ldots, \bm{X}^{(\bm{m}\bm{j})}_{\bm{T}_{\bm{m}\bm{j}}} \mid \bm{I}\right), \\[10pt]
\textbf{where,} & \\
\bm{M} &\text{ is the number of generated sentences,} \\
\bm{T}_{\bm{m}\bm{j}} &\text{ is the length (number of tokens) of sentence } m \text{ obtained } \\&\text{from VLM after }  \bm{j} \text{ fixations, and} \\
\bm{X}^{(\bm{m}\bm{j})}_{\bm{t}} &\text{ is the } t^{\text{th}} \text{ token in sentence } m \text{ obtained from VLM after }\\ &\bm{j} \text{ model fixations.}
\end{aligned}
\right.
\end{equation}

\hfill

\noindent Similar to the cosine similarity reward, we compute the reward at each fixation based on whether the current fixation reduces the model's uncertainty beyond what has already been achieved by previous fixations. Specifically, we compare the average entropy at the current fixation to the minimum average entropy obtained across all earlier fixations. A positive reward is obtained only when the current fixation leads to a new minimum entropy, i.e., when it reduces uncertainty more than any previous fixation. If the current fixation does not improve upon the best (lowest) entropy observed so far, the reward becomes negative and is clipped to zero. Unlike the cosine similarity reward, this formulation does not rely on unfoveated VLM descriptions. Instead, it more directly captures the idea that scene exploration should be rewarded when it progressively reduces uncertainty about what is being seen. Therefore,

\hfill

\begin{equation}
\left.
\begin{aligned}
\bm{R_e}(\bm{j}) &= \max\left(0,\min_{\bm{k}<\bm{j}}\big(\bar{\bm{H}}(\bm{k})\big) - \bar{\bm{H}}(\bm{j})\right), \qquad \bm{j} \geq 1, \\[6pt]
\bm{R_e}(0) &= 0,  \qquad \text{initial fixation}\\[10pt]
\textbf{where,} & \\
\bar{\bm{H}}(\bm{j}) &\text{ denotes the average entropy after } \bm{j} \text{ model fixations.}& \\
\bm{R_e}(\bm{j}) &\text{ denotes the entropy reward for choosing } \bm{j^{th}} \text{  model fixation.}
\end{aligned}
\right.
\end{equation}

\hfill

\subsubsection*{REINFORCE Algorithm}
The fixation prediction problem can be formulated as a reinforcement learning problem in which the eye movement model represents a stochastic policy that selects fixation locations on the image. After each fixation, the model predicts a probability distribution over all possible image patches. Sampling from this distribution corresponds to selecting the next fixation location.

\hfill

\noindent Given that we train separate models for each fixation step, let $\bm{\pi_{\phi_{j}}(a_{j}|s_{j-1})}$ denote the probability of selecting the current fixation location $\bm{a_{j}}$ given the previous state $\bm{s_{j-1}}$, where $\bm{\phi_{j}}$ represents the parameters of the eye movement model for predicting $\bm{j^{\text{th}}}$ model fixation. The state $\bm{s_{j-1}}$ corresponds to the foveated image after $\bm{j-1}$ model fixations. The goal of training is to maximize the expected reward associated with the fixation actions.

\hfill

\noindent Using the REINFORCE algorithm \cite{williams1992simple}, the gradient of the expected reward with respect to the policy parameters can be written as:

\hfill

\begin{equation}
\left.
\begin{aligned}
\nabla_{\bm{\phi}_{\bm{j}}} \bm{J}(\bm{\phi}_{\bm{j}})
&=
\mathbb{E}
\left[
\bm{R}(\bm{j})\nabla_{\bm{\phi}_{\bm{j}}}\log \bm{\pi}_{\bm{\phi}_{\bm{j}}}(\bm{a}_{\bm{j}}|\bm{s}_{\bm{j-1}})
\right], \\[10pt]
\textbf{where,} & \\
\bm{J}(\bm{\phi}_{\bm{j}}) &\text{ is the expected reward objective for the } \\ &\text{ policy choosing } \bm{j^{th}} \text{ model fixation}.
\end{aligned}
\right.
\end{equation}

\hfill

\noindent In practice, this expectation is approximated using a batch of sampled fixation trajectories from $\bm{B}$ images. To reduce the variance of the gradient estimate, we subtract a baseline corresponding to the average reward across the batch:

\hfill

\begin{equation}
\left.
\begin{aligned}
\nabla_{\bm{\phi}_{\bm{j}}} \bm{J}(\bm{\phi}_{\bm{j}})
&=
\frac{1}{\bm{B}}
\sum_{\bm{i}=1}^{\bm{B}}
\left(
\bm{R}_{\bm{i}}(\bm{j}) - \bar{\bm{R}}(\bm{j})
\right)
\nabla_{\bm{\phi}_{\bm{j}}}
\log \bm{\pi}_{\bm{\phi}_{\bm{j}}}(\bm{a}_{\bm{i}\bm{j}}|\bm{s}_{\bm{i}\bm{j-1}}), \\[10pt]
\bar{\bm{R}}(\bm{j})
&=
\frac{1}{\bm{B}}
\sum_{\bm{i}=1}^{\bm{B}}
\bm{R}_{\bm{i}}(\bm{j}), \\[10pt]
\textbf{where,} & \\
\bm{R}_{\bm{i}}(\bm{j}) &\text{ is the reward for choosing image } \bm{j^{th}} \text{ model fixation for image } \bm{i}, \\
\bm{a}_{\bm{i}\bm{j}} &\text{ is the sampled fixation } \bm{j} \text{ for image } \bm{i}, \\
\bm{s}_{\bm{i}\bm{j-1}} &\text{ is the image $\bm{i}$ foveated at the initial and $\bm{j-1}$ previous} \\ &\text{model fixation locations} \\
\bar{\bm{R}}(\bm{j}) &\text{ is the average reward across the batch for choosing fixation } \bm{j}.
\end{aligned}
\right.
\end{equation}

\hfill

\noindent Since fixation locations correspond to spatial positions on the image grid, nearby patches often reveal similar visual information. To encourage spatially smooth learning, we apply Gaussian smoothing to the policy gradient. Instead of assigning the gradient update only to the sampled fixation location, the update is distributed to neighboring patches according to a spatial weighting function. Note that the weights are not normalized. The Gaussian kernel is used purely as a spatial decay function.

\hfill

\noindent Let $\bm{k}$ denote a patch index and $\bm{x_k}$ the spatial coordinate of that patch. The Gaussian weighting function is defined as:

\hfill

\begin{equation}
\left.
\begin{aligned}
\bm{w}(\bm{a},\bm{k})
&=
\exp\left(-\frac{\|\bm{x}_{\bm{a}}-\bm{x}_{\bm{k}}\|^2}{2\bm{\sigma}^2}\right), \\[10pt]
\textbf{where,} & \\
\bm{x}_{\bm{a}} &\text{ is the spatial coordinate of the sampled fixation location } a, \\
\bm{x}_{\bm{k}} &\text{ is the spatial coordinate of patch } k, \text{ and} \\
\bm{\sigma} &\text{ controls the spatial spread of the smoothing kernel.}
\end{aligned}
\right.
\end{equation}

\hfill

\noindent Using this weighting, the smoothed policy gradient becomes:

\hfill

\begin{equation}
\left.
\begin{aligned}
\nabla_{\bm{\phi}_{\bm{j}}} \bm{J}_{\text{smooth}}(\bm{\phi}_{\bm{j}})
&=
\frac{1}{\bm{B}}
\sum_{\bm{i}=1}^{\bm{B}}
\sum_{\bm{k}}
\bm{w}(\bm{a}_{\bm{i}\bm{j}},\bm{k})
\left(
\bm{R}_{\bm{i}}(\bm{j}) - \bar{\bm{R}}(\bm{j})
\right)
\nabla_{\bm{\phi}_{\bm{j}}}
\log \bm{\pi}_{\bm{\phi}_{\bm{j}}}(\bm{k}|\bm{s}_{\bm{i}\bm{j-1}}), \\[10pt]
\textbf{where,} & \\
\bm{w}(\bm{a}_{\bm{i}\bm{j}},\bm{k}) &\text{ distributes the update from sampled fixation } a_{ij} \\
&\text{ to neighboring patch } k.
\end{aligned}
\right.
\end{equation}

\hfill

\noindent This smoothing distributes the gradient update to nearby spatial locations, reflecting the fact that neighboring patches often contain similar information relevant for scene understanding. In practice, this leads to more stable learning and encourages the model to learn coherent spatial fixation policies. Finally, the weights of the model corresponding to each fixation are updated to maximize reward by performing a gradient ascent using the AdamW optimizer, as shown
\begin{equation}
\begin{aligned}
\bm{g}_{\bm{t}} &= \nabla_{\bm{\phi}_{\bm{j}}} \bm{J}(\bm{\phi}_{\bm{j}}), \\[6pt]
\bm{m}_{\bm{t}} &= \bm{\beta}_1 \bm{m}_{\bm{t}-1} + (1-\bm{\beta}_1) \bm{g}_{\bm{t}}, \\[6pt]
\bm{v}_{\bm{t}} &= \bm{\beta}_2 \bm{v}_{\bm{t}-1} + (1-\bm{\beta}_2) \bm{g}_{\bm{t}}^2, \\[6pt]
\hat{\bm{m}}_{\bm{t}} &= \frac{\bm{m}_{\bm{t}}}{1-\bm{\beta}_1^{\bm{t}}}, \\[6pt]
\hat{\bm{v}}_{\bm{t}} &= \frac{\bm{v}_{\bm{t}}}{1-\bm{\beta}_2^{\bm{t}}}, \\[6pt]
\bm{\phi}_{\bm{j}}^{(\bm{t}+1)} &= \bm{\phi}_{\bm{j}}^{(\bm{t})} + \bm{\eta} \frac{\hat{\bm{m}}_{\bm{t}}}{\sqrt{\hat{\bm{v}}_{\bm{t}}}+\bm{\epsilon}} - \bm{\eta} \bm{\lambda} \bm{\phi}_{\bm{j}}^{(\bm{t})},
\end{aligned}
\end{equation}

\begin{align*}
\textbf{where,} \\
\bm{g}_{\bm{t}} &\text{ is the gradient of the objective at iteration } t,\\
\bm{m}_{\bm{t}} &\text{ and } \bm{v}_{\bm{t}} \text{ are the first and second moment estimates,}\\
\bm{\beta}_1,\bm{\beta}_2 &\text{ are the exponential decay rates for these moments,}\\
\bm{\eta} &\text{ is the learning rate,}\\
\bm{\lambda} &\text{ is the weight decay coefficient, and}\\
\bm{\epsilon} &\text{ is a small constant for numerical stability.}
\end{align*}

\hfill

\subsection*{Implementation details}
\subsubsection*{Stimuli}

To enable valid comparisons between human and model eye movements, it is necessary to ensure that the images used by the model correspond to approximately the same visual angles as the images shown to participants during the eye-tracking study. To achieve this, the model's field of view was set to a square region of size $26.6^\circ \times 26.6^\circ$ of visual angle, corresponding to a resolution of $1280 \times 1280$ pixels at an observer distance of 75 cms. All images were embedded within this space at the appropriate size and spatial location to match the presentation conditions used in the eye-tracking experiment. Unless stated otherwise, all analyses described in this Methods section are performed on images embedded within this standardized space.

\hfill

\noindent The training dataset consists of 24,000 images compiled from three publicly available datasets (as described above). From each dataset, a subset of 8,000 images was selected to construct the final training set. Our goal was to ensure that the selected images required more than the initial fixation for the model to understand the scene. To approximate this criterion, we computed the cosine similarity between the descriptions generated from foveated and unfoveated versions of each image. For this comparison, the foveation was placed at the corner of the image. This ensured that the initially revealed region was unlikely to correspond to an informative part of the scene, thereby avoiding the possibility of accidentally selecting a location that might already contain critical information for that specific image. By using a corner location, we ensured that the revealed region was consistently uninformative across images.

\hfill

\noindent Images were then ranked by cosine similarity, and we selected the 8,000 images with the lowest similarity values from each dataset. These images correspond to scenes in which the model's understanding improves substantially when additional regions are revealed, thereby encouraging the model to perform exploratory fixations.

\subsubsection*{Hardware and Software}

Model training was performed using seven NVIDIA RTX A6000 GPUs, each with 48 GB of memory. Since the VLM (OVIS2.5) is used only for inference during training, we employed the vLLM library \cite{kwon2025vllm} to increase inference efficiency. We hosted the VLM on five GPUs and used the remaining two GPUs as client nodes to request outputs from the hosted VLM instances. These two GPUs were responsible for training the eye movement model. Training was performed synchronously across the two GPUs using PyTorch's Distributed Data Parallel (DDP). We modified the vLLM implementation to additionally output the hidden states and token probabilities needed to train the eye movement model.

\hfill


\subsubsection*{Description Sampling from VLM using vLLM library}
\noindent Given that we generate $\bm{M}$ descriptions from the VLM, a common approach would be to use beam search to obtain the top candidate descriptions. However, beam search often produces sentences that follow very similar token trajectories, resulting in highly similar descriptions. This behavior is not ideal in our case because we wish to capture the uncertainty introduced by foveation in the image. To obtain more diverse yet deterministic descriptions, we modified the sampling procedure in vLLM. Specifically, we perform $\bm{M}$ separate greedy decoding runs, each seeded with one of the top-M first tokens. This produces $\bm{M}$ descriptions that differ in their initial token while remaining deterministic during generation, ensuring each description is obtained from a unique trajectory. This sampling method for descriptions was used for all training and analysis done as part of this study.
\hfill

\subsubsection*{Foveation Parameters Calibration}
To enable fair comparison with humans, we aimed to set the foveation parameters (discussed above) to approximate properties of human vision. An observer distance of 75 cm sets the pixels-per-degree value ($\bm{P}$) to 44 pixels/deg, matching our human studies. A Gaussian pyramid of depth 10 was sufficient to cover the resolution drop at any arbitrary distance from the fixation point within the visual field ($26.6^\circ \times 26.6^\circ$).

\hfill

\noindent To match the resolution map of the human eye, we adjusted the $\bm{\alpha}$ value using a separately conducted limited-fixation (two or four) scene description study on humans ($\bm{N}=20$) with 277 images (86 overlapping with the free-viewing dataset). The participant's eyes were tracked using the same setup as in the free-viewing study. A gaze-contingent display was implemented to ensure the image presentation stopped after four fixations for each participant. Since the VLM (OVIS2.5) may not describe all images as accurately as humans with unlimited viewing time, we first selected a subset of 124 images in which the average similarity of VLM descriptions (without foveation) matched the within-human description similarity (unlimited viewing time).

\hfill

\noindent Using this subset, we chose $\bm{\alpha}$ such that the cosine similarity between VLM descriptions of foveated images (based on human fixations from the scene description study) and human descriptions with unlimited viewing time matched the cosine similarity between human descriptions after two or four fixations and those with unlimited viewing time. The resulting value of $\bm{\alpha}$ was $0.63$.

\hfill

\noindent Although participants in our study viewed stimuli from an observer distance of 75 cm, human visual expertise is honed through real-world exposure to objects and faces at specific visual angles. To ensure that our models were trained on ecologically valid scales, we referenced naturalistic observation data from Oruç et al. \cite{oruc2019adult}. Their findings indicate that while the median face size in the adult "face-diet" is approximately $6.0-DVA$, unfamiliar faces, which more closely resemble our test stimuli, subtend a median visual angle of $4.9-DVA$.

\hfill

\noindent In our training dataset, a sample of 1000 images containing people yielded an average face size of only $2.6-DVA$ at the initial 75 cm observer distance. To align these stimuli with the median $4.9-DVA$ value encountered in daily life, we recomputed the required observer distance ($\bm{D}$) to be 39.3 cm. Consequently, the models were trained using the human-calibrated foveation parameter ($\bm{\alpha} = 0.63$), but at this adjusted observer distance, to better approximate the high-resolution features characteristic of natural human social interaction. Testing was still conducted at an observer distance of 75 cm to match human experiments.

\subsubsection*{Action space for the models}

Given a fixed field of view of $1280 \times 1280$ pixels, the action space of the eye movement model consists of 400 possible locations ($1280^2 / 64^2$, see Methods), arranged as a $20 \times 20$ grid. Each grid cell corresponds to a square region of $64 \times 64$ pixels or $1.4 \times 1.4-DVA$ overlaid on the image.

\hfill

\noindent An action $\bm{(X, Y)}$ corresponds to selecting the center of a grid cell as the fixation location. Thus, $(\bm{X}=0, \bm{Y}=0)$ corresponds to the center of the top-left cell, while  $(\bm{X}=19, \bm{Y}=19)$ corresponds to the center of the bottom-right cell.

\subsubsection*{Training procedure}

The eye movement model is trained to generate four sequential fixations following an initial fixation, and each model is trained for a total of five epochs. The effective batch size across two GPUs is 60 images. For each batch, the initial fixation is randomly selected from five predefined locations: the four corners of the image and a location 300 pixels below the center of the field of view. With the current setup, it took 4 days to train a model. To ensure reproducibility, we trained five independent instances of the fRL-SU model and report fixation frequencies and metrics averaged across these instances. Control models (fRL-SC, fRL-Search, and fRL-Entropy) were each trained as a single instance.

\hfill

\noindent Since each fixation is modeled by a separate but architecturally identical Convolutional Neural Network (CNN), the models are trained sequentially. Specifically, each CNN model is trained on five batches (300 images). During training to predict the $\bm{j^{th}}$ fixation, the corresponding CNN model receives the intermediate visual output from the VLM, conditioned on the input image and the previous $\bm{j-1}$ model fixations. The CNN outputs are then transformed into a softmax distribution with temperature $\bm{\tau} = 3$, from which the $\bm{j^{th}}$ fixation is sampled. The preceding $\bm{j-1}$ fixations are obtained via greedy selection from their corresponding frozen CNN models.

\noindent The learning rate ($\bm{\eta}$) for the AdamW optimizer is set to $0.0002$, with weight decay ($\bm{\lambda}$) set to $0.0001$, and all other hyperparameters use the default values provided by the PyTorch implementation of AdamW. The spatial smoothing Gaussian applied in the loss function has a standard deviation of $\bm{\sigma} = 1.5$.

\subsection*{Control Tasks}
To train the eye movement model, the input prompt to the VLM is fixed, instructing it to describe the given input image. As a control, we also train the eye movement model using alternative input prompts corresponding to different tasks and compare the resulting eye movements with human eye movements. Specifically, we trained two additional models: one performing a search task and another performing a scene categorization task based on categories from Anderson et al. \cite{anderson2021category}. We use the following prompts for the respective tasks.

\begin{center}
\begin{minipage}{0.85\linewidth}
\textbf{Text prompt 1 (Search task):}

\textit{``Provide your best description of the smallest man-made object present and clearly visible in the image in a sentence. Do not mention the blur seen in the picture.''}
\end{minipage}
\end{center}

\begin{center}
\begin{minipage}{0.85\linewidth}
\textbf{Text prompt 2 (Scene Classification task):}

\textit{``Analyze this scene and classify it by selecting exactly one label from each of the three dimensions below. Do not use any outside vocabulary.\
Semantic Content: Nature, Road, Residence, Farm, Beach, Car Park\
3D Spatial Structure: Cluttered, Closed Off, Flat, Tunnel\
2D Image Appearance: Dark, Bright, Blue, Green, Brown\
Format your output strictly as:\
Semantic: [label]\
Spatial: [label]\
Appearance: [label]''}
\end{minipage}
\end{center}

\noindent For simplicity, the models trained using the above prompts were optimized solely using the semantic accuracy reward. Additionally, we trained models with very high ($\bm{\alpha = 0.2}$) and very low foveation ($\bm{\alpha = 20}$). All models were trained, at an observer distance of 39.3 cm, to make four fixations after the initial fixation (see details above).

\subsection*{Existing fixation prediction models}

\subsubsection*{DeepGaze}

DeepGaze \cite{linardos2021deepgaze} is a neural network model that combines low-level image features from a ResNet-50 CNN and human fixation data to predict a fixation density map. This study uses an implementation provided by Kümmerer \cite{DeepGaze}. Since this results in a static map for each image, we adapt it to generate sequential fixations in our test dataset by smoothing the DeepGaze map with a $2-DVA$ circular average kernel and selecting four fixation locations sequentially using inhibition of return (IOR).

\hfill

\noindent The IOR is modeled as a circular patch whose size is set to match the average saccade distance per fixation observed in human free-viewing data. For DeepGaze, the diameter of the IOR patch is $2.6-DVA$.

\subsubsection*{Saliency Models}

As part of this study, we compare two low-level saliency models: Itti Koch \cite{itti_model_1998} and Graph-Based Visual Saliency (GBVS) \cite{harel2006graph}. The Itti-Koch model predicts fixation locations by combining multi-scale feature contrasts (color, intensity, orientation) into a bottom-up saliency map. GBVS extracts similar low-level features but constructs a graph over them and uses Markov chains to generate a saliency map. This study uses an implementation provided by Kümmerer \cite{PySaliency}.

\hfill

\noindent Similar to DeepGaze, we generate sequential fixations by selecting four locations per image through iterative maximum selection on smoothed ($2-DVA$ kernel) saliency maps with inhibition of return matched to human data. For GBVS, the IOR diameter is $2.8-DVA$, while for the Itti-Koch model, it is $2-DVA$. The differences in IOR diameters reflect the sparsity of the predicted saliency maps, with sparser maps requiring smaller IOR regions.

\subsubsection*{Random}

For the random baseline, 1000 samples of four fixations were uniformly drawn from the pixels corresponding to the image region within the field of view. The resulting average inter-fixation distance was substantially larger ($11-DVA$) than that observed in human data, even without applying inhibition of return. Nevertheless, we imposed an IOR constraint with a diameter of $2-DVA$ during sampling.

\subsection*{Fixation prediction heatmaps for trained models}
The eye movement model outputs a probability map of size $20 \times 20$ for each predicted fixation; thus, four fixations yield four such maps. To obtain a single static fixation prediction map, we compute the element-wise maximum across these maps. This operation aggregates the most probable regions from each fixation step, assuming that each map highlights distinct regions of interest. Taking the maximum, therefore, combines these high-probability regions into a single map, providing a consolidated estimate of likely fixation locations.

\subsection*{Measured fixations heatmaps}\label{Methods:heatmapfix}

Fixations on an image were represented as pixel-wise white dots on a blank canvas and convolved with a Gaussian kernel with a standard deviation of $0.25^\circ$ visual angle. OpenCV2's \texttt{GaussianBlur} function was used for this purpose. The convolved fixation points were superimposed onto a blank map at their corresponding $\bm{(x, y)}$ locations to generate the heatmap. To match the trained models, the heatmaps were generated for the first four measured fixations after the initial fixation.

\subsection*{Fixation prediction heatmap comparison metrics}

In this study, we use the following standard metrics from the literature: Area Under ROC (AUC) and Correlation Coefficient.

\begin{itemize}
    \item \textbf{AUC:} Measures how well a saliency map discriminates fixation locations from non-fixated locations.
    \item \textbf{CC:} Measures the linear similarity between the predicted saliency map and the human fixation density map.
\end{itemize}

This study uses an implementation of these metrics provided by Kümmerer \cite{PySaliency}.

\subsection*{Fixation frequency to categories comparisons}

\begin{figure}[htbp]
    \centering
     \includegraphics[width=\textwidth]{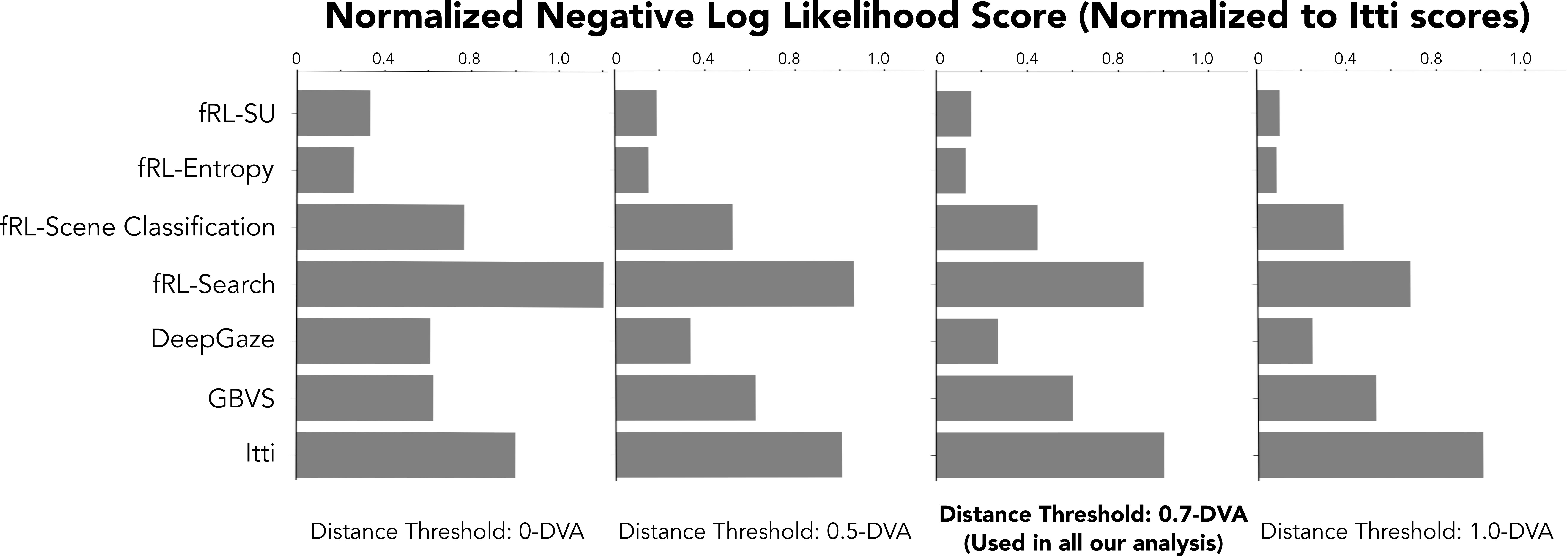} 
    \caption{Model-human fixation frequency agreement across segmentation distance thresholds. Normalized Negative Log Likelihood (NNLL) scores, normalized to Itti-Koch Saliency scores, quantifying how closely each model’s fixation frequencies match those of human participants. Lower values indicate closer agreement. Each panel corresponds to a different distance threshold used to assign a fixation to a scene element. Across all thresholds, fRL-SU and fRL-Entropy show the closest agreement with human data, while fRL-Search and Itti-Koch show the largest deviations. The relative ordering of the models is the same regardless of the distance threshold, indicating that the main findings are not an artifact of the $0.7-DVA$ criterion. }
    \label{fig:S1}
\end{figure}

To compare fixation patterns, we classified the visual elements in each scene and quantified how frequently the first four subsequent fixations fell within these categories. We generated binary segments for each object or region, using a spatial tolerance of $0.7-DVA$ from the segment edges to determine category membership (see fig.\ref{fig:S1} for generalization of results with other tolerances). This analysis was applied across humans, saliency models, and our trained and control models. The categories are as follows:

\subsubsection*{People:}
This category corresponds to all people present in an image. The Segment Anything Model 3 (SAM3) \cite{carion2025sam} was used to detect and segment people in each scene, and the resulting masks were used. Of the 147 test images, 85 contain people.

\subsubsection*{Center bias:}
This corresponds to the central $5-DVA$ region in each image. A circular binary mask of size $5-DVA$ centered in each image was used to count fixations falling into this category. All 147 images in the dataset include a center bias mask.

\subsubsection*{SU-R Object:}
This corresponds to objects that are critical for understanding the given image. Based on previous work on scene understanding maps (SUMs) \cite{murlidaran2025eye, murlidaran2025semantic}, these critical objects were identified by quantifying the impact of removing objects/people from a scene on the scene descriptions generated by a Multi-Modal Large Language Model (MLLM). The object/people removal, scene description generation, and impact measurement using embedding-based description ratings were all automated (AUTOSUM), as detailed in \cite{murlidaran2025semantic}. To ensure consistency with the SU-R Object criteria, we excluded any people present in the scene from the AUTOSUM predictions. The remaining scores were normalized relative to the most impactful object in each image. Segments achieving a normalized score of 0.95 or higher were classified as SU-Relevant (SU-R) Objects; these segments then served as masks for calculating fixation frequency.

\hfill

\noindent The SU-R object category was further split into the following subcategories:

\paragraph{SU-R Object (Gaze/Grasp)}
These are images where the SU-R Object is either grasped or gazed at by people present in the scene. Naturally, this forms a subset of images that contain people. Of the 85 images with people, 53 include gaze or grasp cues directed toward the SU-R Object, and the SU-R masks for these 53 images are included in this category.

\paragraph{SU-R Object (No Gaze/Grasp)}
These are images in which the SU-R Object has no gaze or grasp cues directed toward it. All images without people naturally fall into this category. There are 45 such images that contain SU-R Objects with no gaze/grasp cues, and their masks are included here. We ignore images that contain people but lack gaze/grasp cues toward the SU-R Object due to the limited number of such cases (only 9 images).

\subsubsection*{SU-I Objects:}
All objects with an AUTOSUM score below 0.95 were considered SU-Irrelevant (SU-I) Objects. Since there are typically many SU-I objects in each scene, their masks were normalized by the number of SU-I objects per image to ensure fair comparison with other categories.

\hfill

\noindent Similar to SU-R Objects, SU-I Objects were split using the same set of images as the two SU-R categories (53 and 45 images, respectively), ensuring fair comparisons.

\subsubsection*{Text:}
Images containing text were manually segmented, and 4-sided polygon masks were created for text regions. Of the 147 images, 51 contain text regions.

\subsubsection*{Saliency (GBVS):}
This category corresponds to regions identified by the top prediction of the GBVS saliency map. For images containing an SU-R Object, a circular patch centered at the top prediction location, with an area matching the SU-R Object segment, was used as the mask. For images without an SU-R Object, a circular patch matching the average size of SU-R Objects across the test dataset was used. All 147 images include a saliency mask.

\subsubsection*{Comparison of model and human fixation frequency distributions}
To assess how well the models predict the distribution of human fixation frequencies across these categories, we measure the divergence between the models' predicted fixation frequencies and the empirical human distribution. We frame this as a Negative Log-Likelihood (NLL) problem. By modeling the human fixation frequencies across object categories as a multivariate Gaussian distribution $\mathcal{N}(\mu, \Sigma)$, we calculate the NLL of the model's predictions under this human prior. 

\hfill

\begin{figure}[htbp]
    \centering
     \includegraphics[width=0.6\textwidth]{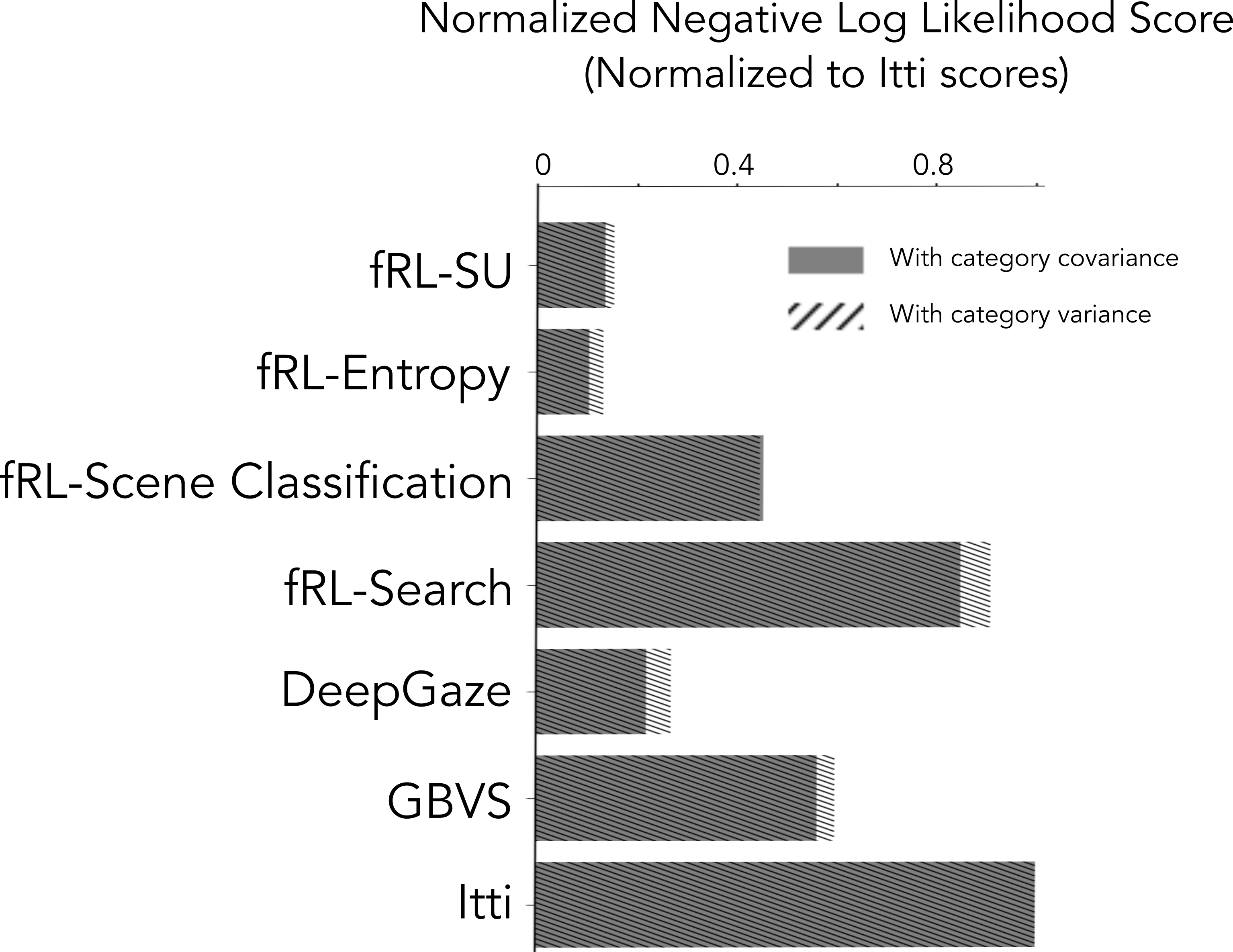} 
    \caption{Category independent vs category dependent NNLL comparisons. NNLL scores, normalized to Itti-Koch, comparing each model’s fixation frequencies to human data. Solid bars show the NNLL computed with category covariance, treating categories as dependent. Hatched bars show the NNLL computed using only category-specific variances, treating categories as independent. The values are similar, indicating that dependencies within the categories minimally affect the NNLL scores.}
    \label{fig:S4}
\end{figure}

\noindent We compute two variants of NLL: one assuming the categories are independent (the main results), and the other assuming they are not (see fig. \ref{fig:S4}). The independent variation is as follows:

\begin{equation}
\begin{aligned}
\textbf{NLL}_{\textbf{indep}} &= -\sum_{\boldsymbol{i}=1}^{\boldsymbol{k}} \left[ \frac{(\boldsymbol{x}_{\boldsymbol{i}} - \boldsymbol{\mu}_{\boldsymbol{i}})^2}{2\textbf{SE}_{\boldsymbol{i}}^2} + \frac{1}{2}\ln(2\pi\textbf{SE}_{\boldsymbol{i}}^2) \right] \\
\textbf{where,} & \\
\boldsymbol{k}  &\text{ is total number of object categories} \\
\boldsymbol{x}_{\boldsymbol{i}} &\text{ is the model's predicted fixation frequency for category } \boldsymbol{i} \\
\boldsymbol{\mu}_{\boldsymbol{i}} &\text{ is the mean empirical human fixation frequency for category } \boldsymbol{i} \\
\textbf{SE}_{\boldsymbol{i}} &\text{ is the standard error of the human frequencies for category } \boldsymbol{i}
\end{aligned}
\end{equation}

\noindent In the dependent variation, we calculate the covariance between categories based on the fixation frequencies of 50 participants in the free-viewing study, averaged across all 147 images in the test dataset. The NLL is computed as follows:

\begin{equation}
\begin{aligned}
\mathbf{NLL}_{\mathbf{mvn}} &= -(\frac{1}{2} (\boldsymbol{x} - \boldsymbol{\mu})^{\boldsymbol{T}} \boldsymbol{\Sigma}^{\boldsymbol{-1}} (\boldsymbol{x} - \boldsymbol{\mu}) + \frac{1}{2} \ln |\boldsymbol{\Sigma}| + \frac{\boldsymbol{k}}{2} \ln(2\boldsymbol{\pi})) \\
\textbf{where,} & \\
\boldsymbol{x} &\text{ is the vector of the model's predicted fixation frequencies across all categories} \\
\boldsymbol{\mu} &\text{ is the vector of the mean empirical human fixation frequencies} \\
\boldsymbol{\Sigma} &\text{ is the covariance matrix of the empirical human data} \\
\boldsymbol{\Sigma}^{\boldsymbol{-1}} &\text{ is the inverse of the covariance matrix (precision matrix)} \\
|\boldsymbol{\Sigma}| &\text{ is the determinant of the covariance matrix} \\
\boldsymbol{T} &\text{ is the transpose operator} \\
\boldsymbol{k} &\text{ is the total number of object categories}
\end{aligned}
\end{equation}

\noindent To allow comparison across datasets, the NLL scores were normalized by the NLL score of the Itti-Koch saliency model, yielding the Normalized NLL (NNLL):

\begin{equation}
\text{NNLL} = \frac{\text{NLL}_{\text{model}}}{\text{NLL}_{\text{Itti-Koch}}}
\tag{S23}
\end{equation}

\noindent A score closer to 0 indicates better agreement with human fixation frequencies, while a score of 1 indicates performance equivalent to the Itti-Koch baseline, the worst-performing model in this study.

\subsection*{Description similarity comparisons}

Given that the VLM generates image descriptions in response to the text prompt (defined above) and that the eye movement model can be constrained to fixate at specified locations, we use fixation sequences from various models and participants in the test dataset and force the eye movement model to follow these sequences. We then record the model-generated descriptions for each fixation sequence.

\hfill

\noindent These descriptions are compared to the unfoveated model descriptions of the same image using cosine similarity between sentence embeddings obtained from the Jasper embedding model \cite{zhang2024jasper}. To compute the description similarities for the random model, we used only 10 out of the 1000 generated samples (see Methods above) to keep the computation tractable.

\subsection*{Testing on publicly available dataset}
To evaluate our models, we use an external dataset from a recent free-viewing study by Linka et al. \cite{linka2025protracted}. This dataset comprises a subset of the OSIE dataset (40 images; OSIE 40, \cite{linka2020osieshort, xu2014predicting}) and includes a substantial amount of participant data ($\bm{N = 6,720}$) collected across different age groups. Linka et al. analyzed these data by calculating the proportion of fixations falling into four distinct categories: 1) Head, 2) Body, 3) Touched (objects actively touched by people present in the scene), and 4) Text.

\subsubsection*{Stimuli Dimensions}
Because observer distance in the Linka et al. \cite{linka2025protracted} study varied between 50 cm and 90 cm, the subtended visual angle of the images was not constant. Therefore, we maintained the original stimulus resolution of $1000 \times 750$ pixels and positioned the images centrally within the models' field of view. We fixed the observer distance to 75 cm to match our testing conditions. Given that the pixel pitch (0.294 mm) of the monitor used in their study (ProLite T1931SR-B5) is nearly identical to ours (0.293 mm), the centrally placed images subtend $21^\circ \times 16^\circ-DVA$ within the models' field of view.

\subsubsection*{Fixation Analysis}
Participants in the Linka et al. \cite{linka2025protracted} study initiated viewing from the screen center. To ensure a fair comparison, the initial fixation for all baseline models (DeepGaze, other saliency models, and the Random baseline), as well as the participants, is set to the screen center. In contrast, our trained models inherently predict the first fixation toward the center, so their starting point is  $10.1^\circ$ below the center of the field of view, consistent with our primary analyses. For all comparative analyses, these respective initial starting points are treated as the "first fixation" across all models and participants. Consistent with the methodology used for our primary dataset, we evaluate the first four fixations. Furthermore, we compute fixation frequencies—rather than proportions—for both the four original categories from Linka et al. For our categories, we use data from adults only (ages 18-72) and the SU-R Object and SU-I Object categories derived from AutoSUM. We also calculate fixation frequencies at the top-saliency location predicted by the GBVS model. Finally, we compute the cosine similarity across the first four fixations for all models and participants.

\subsection*{Bootstrap Analysis}
Error bars for fixations to categories were obtained using 10000 bootstrap resampling of participants. The images were not bootstrapped as doing so would change the number of images in each category. For the AUCs and correlation coefficients, error bars were obtained using 1000 bootstrap resampling of participants and images.


\subsection*{Supplementary Tables and Figures}

\begin{figure}[htbp]
    \centering
     \includegraphics[width=0.7\textwidth]{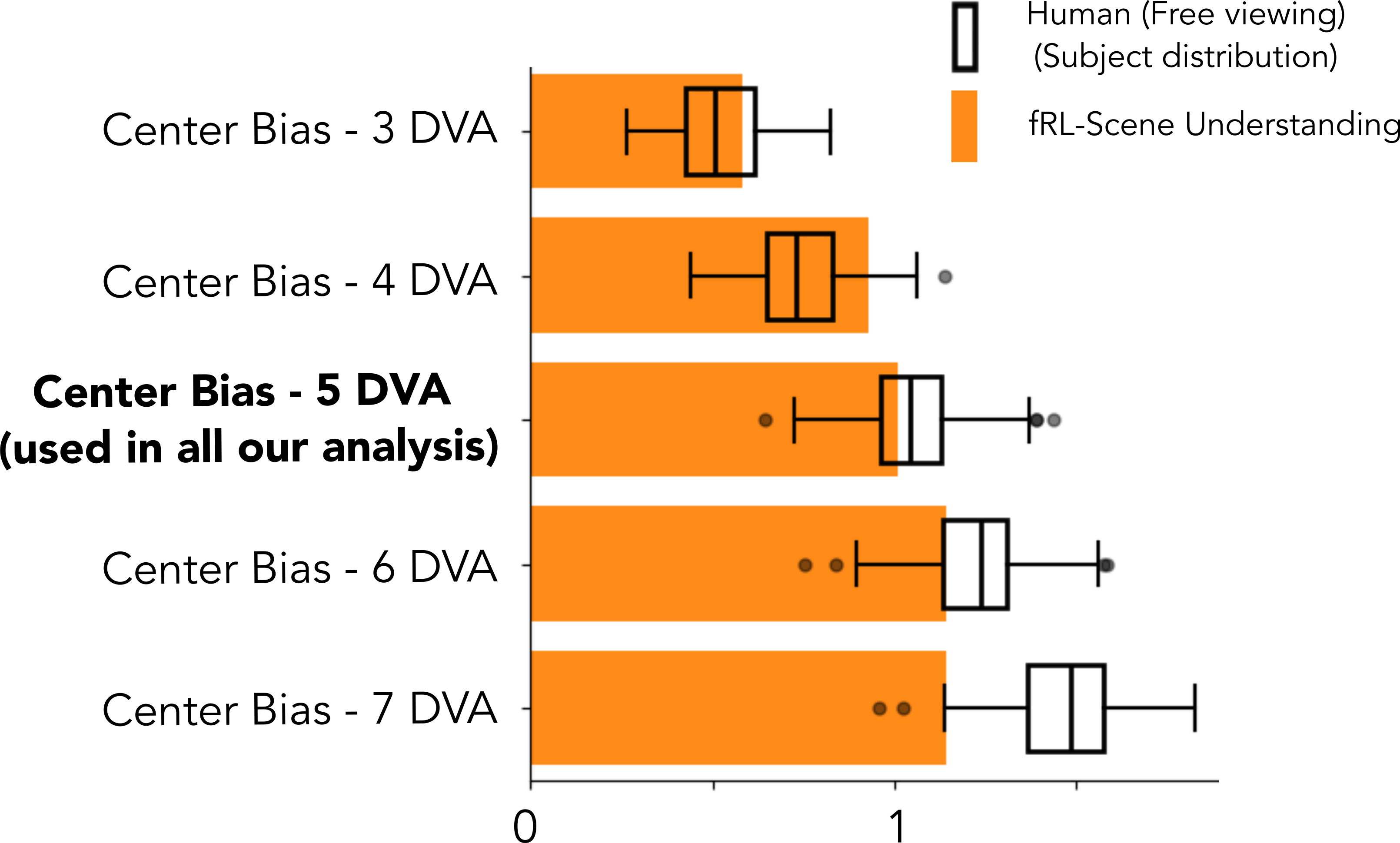} 
    \caption{Center bias fixation frequency across various degrees of visual angle (DVA). Fixation frequency within the central region of the image for human free-viewing participants (box plots) and the fRL-SU model (orange bars), computed using circular central regions of 3,4,5,6 and 7-DVA. The DVA size of 5, used in all main analyses, was selected based on prior human fixation data. At this size, fRL-SU central fixation frequency falls within the human subject distribution.}
    \label{fig:S2}
\end{figure}

\begin{figure}[htbp]
    \centering
     \includegraphics[width=0.7\textwidth]{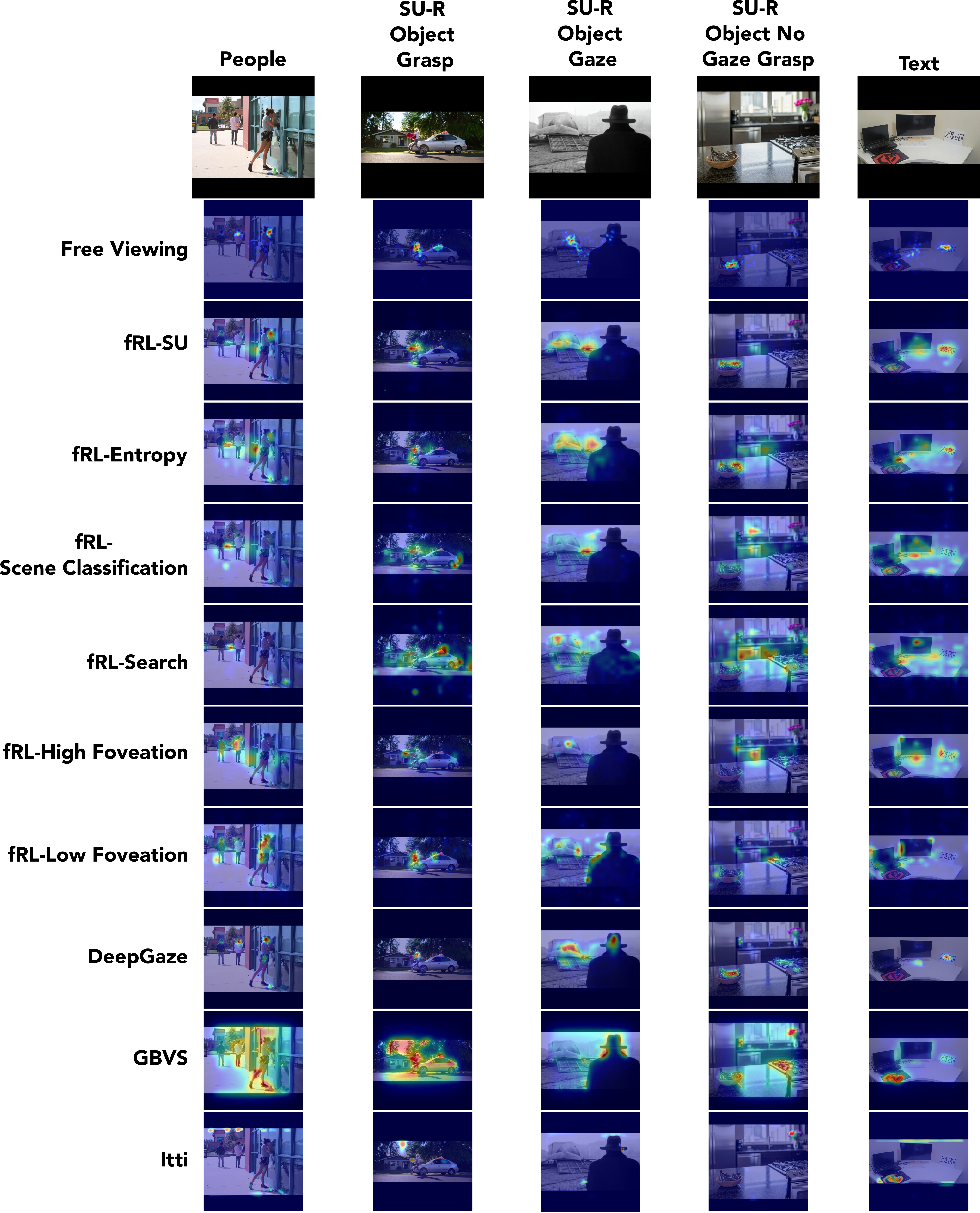} 
    \caption{Fixation heatmap examples across models and scene element categories. Example images from four scene element categories are shown in columns: scenes with people, scenes with SU-relevant objects gazed or grasped, scenes with SU-relevant objects without gaze or grasp, and scenes containing text. Rows show fixation heatmaps generated by human free-viewing participants and each computational model. }
    \label{fig:S3}
\end{figure}

\begin{figure}[htbp]
    \centering
     \includegraphics[width=\textwidth]{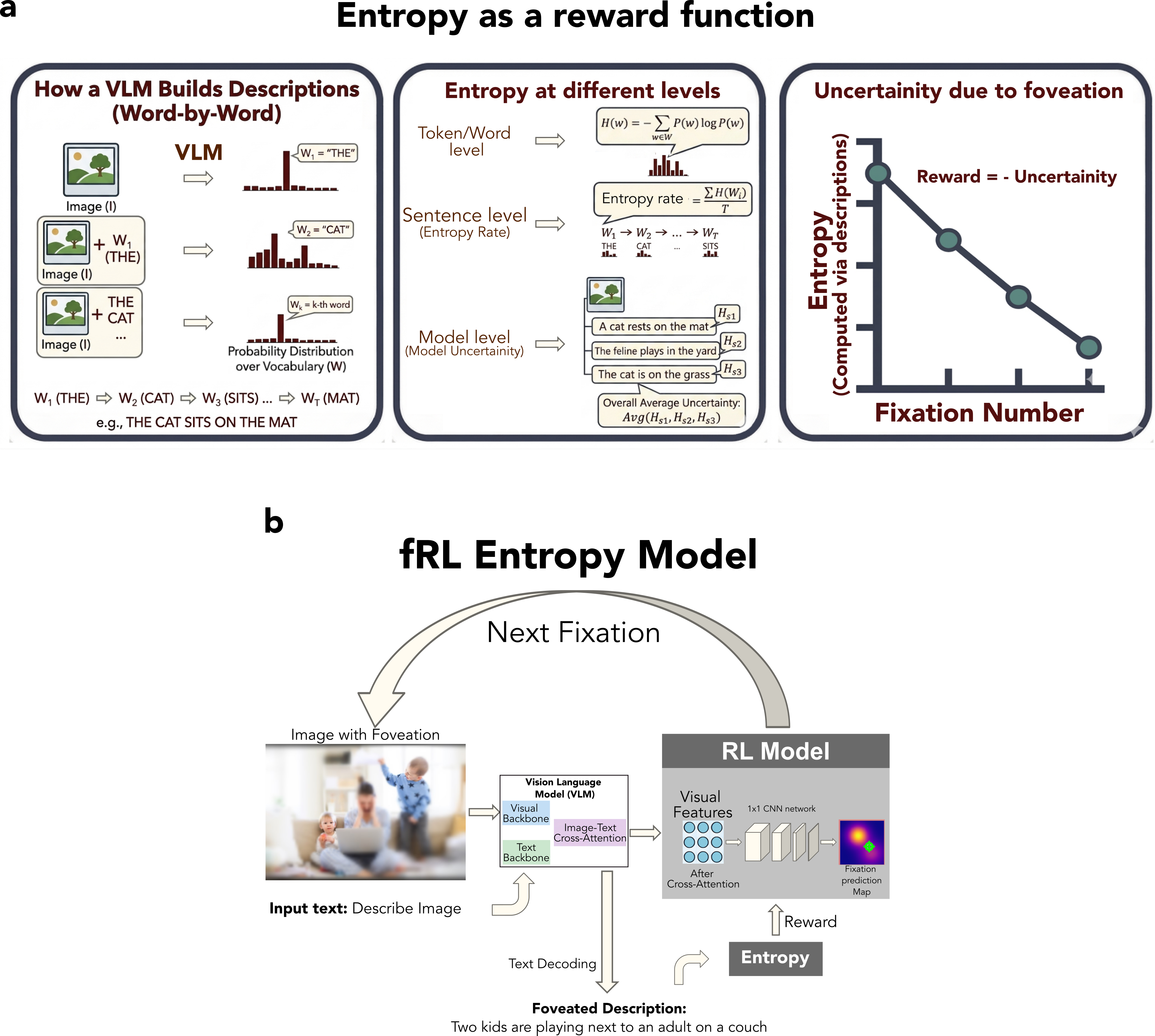} 
    \caption{fRL-Entropy model architecture and training pipeline. (a) Entropy as a reward signal. A VLM generates descriptions word-by-word, producing a probability distribution at each word. Token-level entropy is averaged across tokens to obtain a sentence-level entropy score, then averaged across ten candidate descriptions to estimate overall uncertainty. Reward is defined as the negative of this uncertainty. (b) Schematic of the fRL-Entropy model, which shares the fRL-SU architecture but replaces the cosine-similarity reward with the entropy-based reward, removing the need for ground-truth descriptions during training.}
    \label{fig:S5}
\end{figure}

\begin{figure}[htbp]
    \centering
     \includegraphics[width=0.8\textwidth]{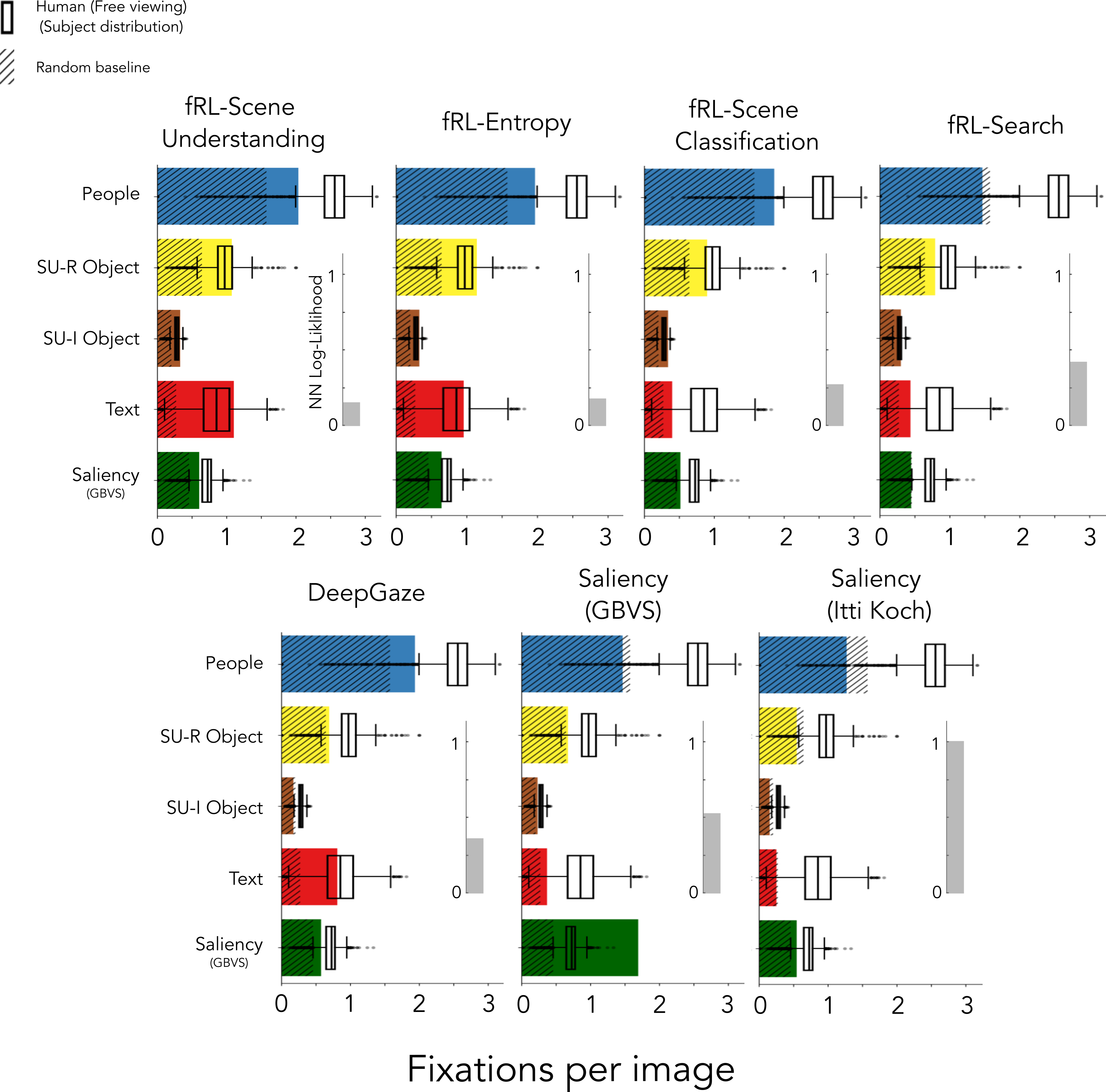} 
    \caption{Generalization to the Linka et al. \cite{linka2025protracted} dataset. Fixation frequencies per image across scene element categories for each model compared to the free-viewing participants in the dataset. Box plots show the human subject distribution; the bars show model predictions; hatched bars show the random baseline. }
    \label{fig:S6}
\end{figure}

\begin{table}[H]
\centering
\begin{tabular}{lccc}
\hline
\textbf{Model} & \textbf{AUC} & \textbf{CC} & \textbf{p-value} \\
\hline
fRL-SU                   & 0.9046 $\pm$ 0.0038 & 0.3715 $\pm$ 0.0121 & -    \\
fRL-Scene Classification & 0.8707 $\pm$ 0.0049 & 0.2785 $\pm$ 0.0090 & *** \\
fRL-Search               & 0.8088 $\pm$ 0.0061 & 0.1832 $\pm$ 0.0097 & *** \\
DeepGaze                 & \textbf{0.9312} $\pm$ 0.0036 & \textbf{0.6281} $\pm$ 0.0164 & *** \\
GBVS                     & 0.8451 $\pm$ 0.0051 & 0.2382 $\pm$ 0.0086 & *** \\
Itti Koch                & 0.5329 $\pm$ 0.0048 & 0.0596 $\pm$ 0.0093 & *** \\
\hline
\end{tabular}
\caption{AUROC and correlational coefficient (CC) metrics comparison of different models. fRL-SU is significantly better than other models in comparison, except DeepGaze. DeepGaze is the best-performing model across both metrics.}
\label{tab:model_performance}
\end{table}

\end{document}